\documentclass[acmtog,screen,nonacm]{acmart}

\usepackage{booktabs} 

\usepackage{multirow}
\usepackage{bbm}

\usepackage[ruled]{algorithm2e} 

\SetAlFnt{\small}
\SetAlCapFnt{\small}
\SetAlCapNameFnt{\small}
\SetAlCapHSkip{0pt}

\setcopyright{acmcopyright}\acmJournal{TOG}
\acmYear{2022}\acmVolume{41}\acmNumber{4}\acmArticle{104}\acmMonth{7} \acmDOI{10.1145/3528223.3530108}

\usepackage[inline]{enumitem} 
\usepackage{color}
\usepackage{multirow}
\usepackage[makeroom]{cancel}

\usepackage{caption}
\definecolor{turquoise}{cmyk}{0.65,0,0.1,0.3}
\definecolor{purple}{rgb}{0.65,0,0.65}
\definecolor{dark_purple}{rgb}{0.5,0,0.5}
\definecolor{dark_green}{rgb}{0, 0.5, 0}
\definecolor{orange}{rgb}{0.8, 0.6, 0.2}
\definecolor{red}{rgb}{0.8, 0.2, 0.2}
\definecolor{darkred}{rgb}{0.6, 0.1, 0.05}
\definecolor{blueish}{rgb}{0.0, 0.3, .6}
\definecolor{light_gray}{rgb}{0.7, 0.7, .7}
\definecolor{pink}{rgb}{1, 0, 1}
\definecolor{greyblue}{rgb}{0.25, 0.25, 1}

\renewcommand{\comment}[1]{}

\newcommand{\CIRCLE}[1]{\raisebox{.5pt}{\footnotesize \textcircled{\raisebox{-.6pt}{#1}}}}

\newcommand{\Figure}[1]{Figure~\ref{fig:#1}}

\newcommand{\eq}[1]{\eqref{eq:#1}}

\newcommand{\Section}[1]{Section~\ref{sec:#1}}

\usepackage{blindtext}

\newcommand{\SupplementaryMaterial}{{{\color{dark_purple}supplementary material}}\xspace}
\newcommand{\SupplementaryVideo}{{{\color{dark_purple}supplementary video}}\xspace}

\usepackage{dsfont}
\DeclareMathOperator*{\argmin}{arg\,min}

\newcommand{\Grid}{\mathcal{G}}
\newcommand{\GridEdges}{\mathcal{E}}
\newcommand{\GridVertices}{\mathcal{X}}
\newcommand{\loss}[1]{\mathcal{L}_{#1}}

\newcommand{\real}{\mathbb{R}}
\newcommand{\bool}{\mathbb{B}}
\newcommand{\EdgeVertices}{{\Vertices^\GridEdges}}
\newcommand{\EdgeNormals}{{\mathcal{N}^\GridEdges}}
\newcommand{\Vertices}{\mathcal{V}}
\newcommand{\Faces}{\mathcal{F}}

\newcommand{\Signs}{\mathcal{S}}

\newcommand{\SDF}{\Phi}
\newcommand{\edge}{\mathbf{e}}
\newcommand{\fun}{f}
\newcommand{\given}{;~}
\newcommand{\pars}{\theta}
\newcommand{\NDC}{NDC\xspace}
\newcommand{\UNDC}{UNDC\xspace}
\newcommand{\BCE}{\text{BCE}}
\newcommand{\gt}{\text{gt}}
\newcommand{\Masks}{\mathcal{M}}
\newcommand{\Input}{\mathcal{I}}

\begin{document}
\title{Neural Dual Contouring}

\author{Zhiqin Chen}
\affiliation{
 \institution{Simon Fraser University}
 \country{Canada}}
\email{zhiqinc@sfu.ca}

\author{Andrea Tagliasacchi}
\affiliation{
 \institution{Google Research, Simon Fraser University}
 \country{Canada}}
\email{atagliasacchi@google.com}

\author{Thomas Funkhouser}
\affiliation{
 \institution{Google Research}
 \country{USA}}
\email{tfunkhouser@google.com}

\author{Hao Zhang }
\affiliation{
 \institution{Simon Fraser University}
 \country{Canada}}
\email{haoz@sfu.ca}

\begin{abstract}
We introduce \textit{neural dual contouring} (NDC), a new data-driven approach to mesh reconstruction based on dual contouring (DC).  Like traditional DC, it produces exactly one vertex per grid cell and one quad for each grid edge intersection, a natural and efficient structure for reproducing sharp features.   However, rather than computing vertex locations and edge crossings with hand-crafted functions that depend directly on difficult-to-obtain surface gradients, NDC uses a neural network to predict them.   As a result, NDC can be trained to produce meshes from signed or unsigned distance fields, binary voxel grids, or point clouds (with or without normals); and it can produce open surfaces in cases where the input represents a sheet or partial surface.   During experiments with five prominent datasets, we find that NDC, when trained on one of the datasets, generalizes well to the others. Furthermore, NDC provides better surface reconstruction accuracy, feature preservation, output complexity, triangle quality, and inference time in comparison to previous learned (e.g., neural marching cubes, convolutional occupancy networks) and traditional (e.g., Poisson) methods.
Code and data are available at
\href{https://github.com/czq142857/NDC}{https://github.com/czq142857/NDC}.

\end{abstract}



\begin{teaserfigure}
\centering
\includegraphics[width=1.0\linewidth]{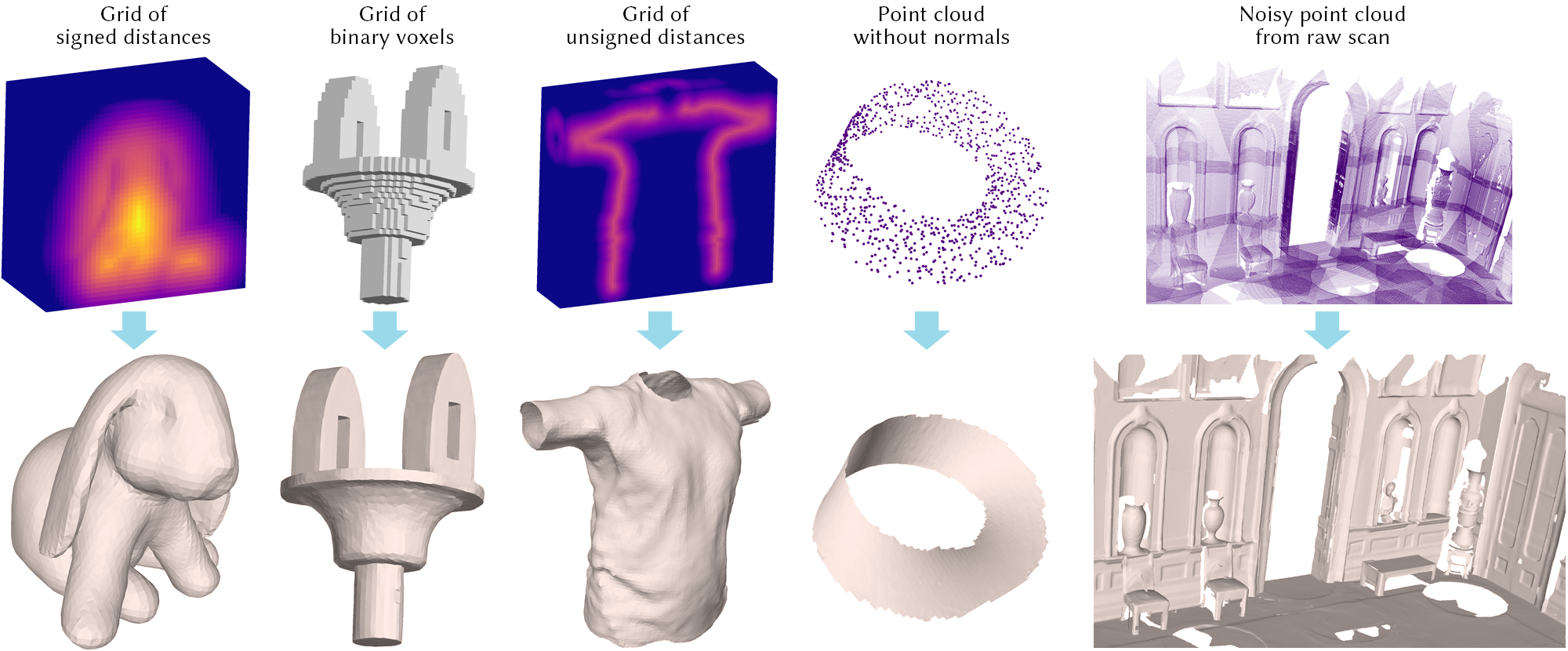}
\caption{Neural dual contouring (NDC) is a unified data-driven approach that learns to reconstruct meshes (bottom) from a variety of inputs (top): signed or unsigned distance fields, binary voxels, non-oriented point clouds, and noisy raw scans. Trained on CAD models, NDC generalizes to a broad range of shape types: CAD models with sharp edges, organic shapes, open surfaces for cloths, scans of indoor scenes, and even the non-orientable Mobi\"{u}s strip.}
\label{fig:teaser}
\end{teaserfigure}

\maketitle

\section{Introduction}
\label{sec:intro}

Polygonal mesh reconstruction from discrete inputs such as point clouds and voxel grids has been one of the most classical
and well-studied problems in computer graphics~\cite{araujo2015survey,berger2017survey}. 
Current solutions to the problem are predominantly model-driven, often relying on assumptions such as those
related to shape characteristics (e.g., watertightness, zero genus, etc.), surface interpolants (e.g., trilinearity), 
sampling conditions, surface normals, and other reconstruction priors.
It is only recently that a few {\em data-driven\/} meshing methods have emerged.  
However, they have mostly focused on learning point set triangulations~\cite{rako2021cvpr,sharp2020eccv,liu2020eccv}.  
One exception is Neural Marching Cubes (NMC) ~\cite{chen2021nmc},
a learning-based Marching Cubes (MC) approach for mesh reconstruction
from a voxel grid of signed distances or binary occupancies. In comparison to
the original MC algorithm~\cite{lorensen1987marching} and its best-known variant, MC33~\cite{chernyaev1995marching},
NMC uses tessellation templates with more adaptive
mesh topologies and learns local shape priors from training meshes.
As a result, NMC generalizes well to a broader range of shape
types and excels at preserving sharp features, two
long-standing issues in existing MC work.
On the other hand, the NMC tessellation templates are necessarily more complex than those of MC and MC33. 
As a result, NMC typically outputs 4-8 times the number of triangles and incurs 100$\times$ or 
more compute time to reconstruct a mesh. 

\begin{figure}
\centering
\includegraphics[width=0.99\linewidth]{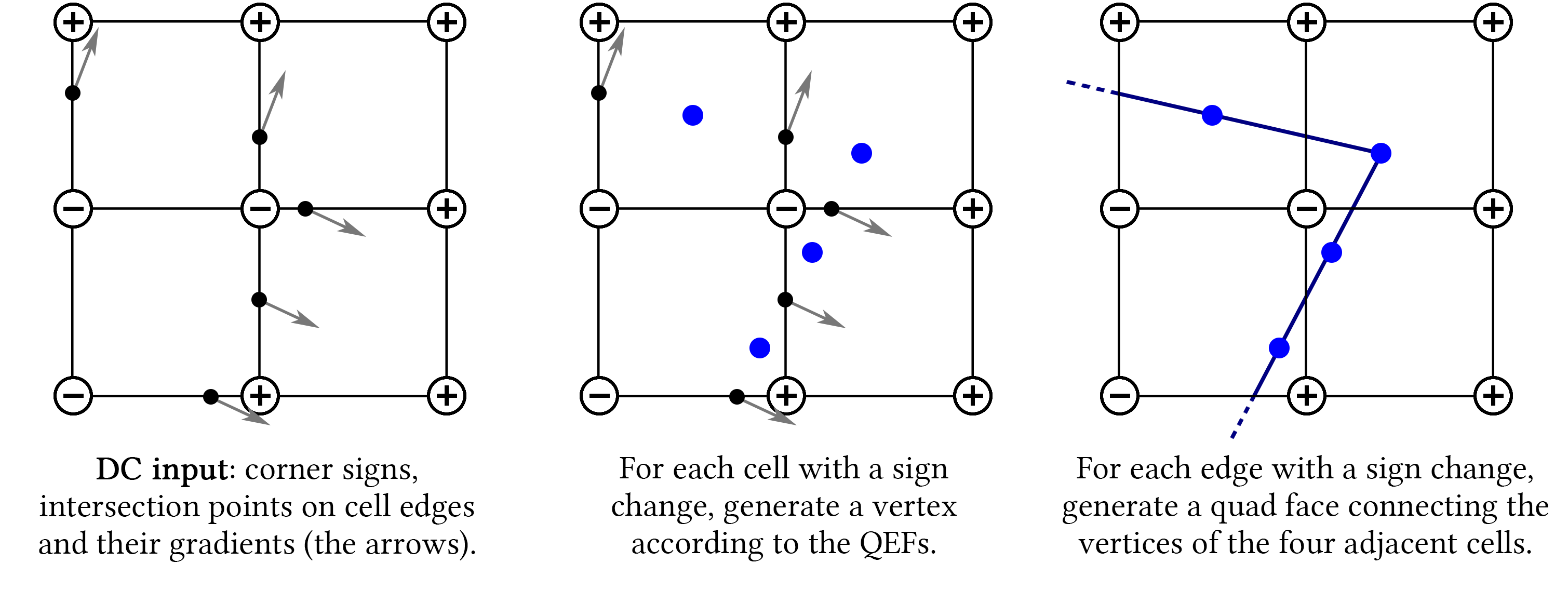}
\includegraphics[width=0.99\linewidth]{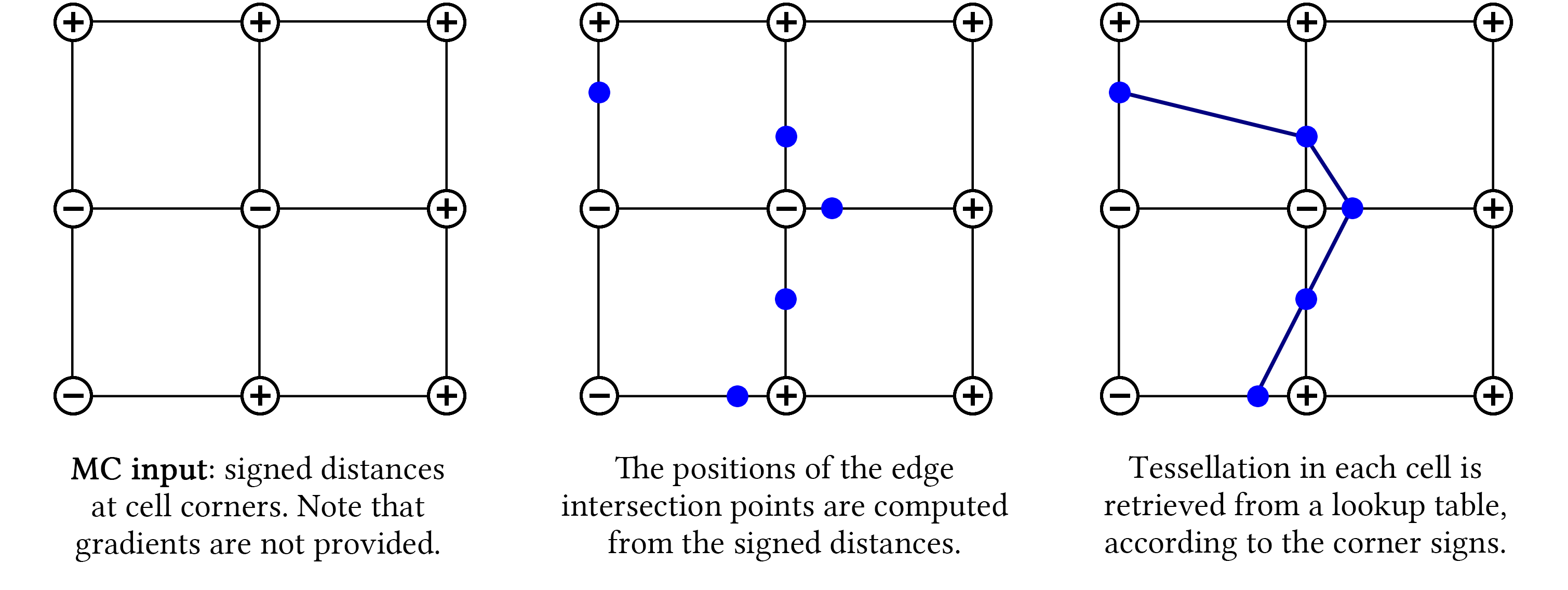}
\caption{\textbf{Dual Contouring (DC)} vs. \textbf{Marching Cubes~(MC)} -- visualized in 2D on different inputs that were sampled from the same underlying shape, 
DC (top) reconstructs a sharp feature (as an intersection between faces, in the top-right cell), while MC (bottom) does not.}
\vspace{-5mm}
\label{fig:mc_vs_dc}
\end{figure}

In this paper, we introduce \textit{Neural Dual Contouring} (NDC), a new data-driven approach to mesh reconstruction 
based on dual contouring (DC) ~\cite{ju2002dual}. The key motivation for building our learning framework upon DC rather than 
MC is that it provides a {\em more natural\/} and {\em more efficient\/} means of reproducing sharp features.
As shown in \Figure{mc_vs_dc}, NDC only needs to predict {\em one\/} mesh vertex per grid cell (i.e., a cube) and one quad for each cell edge intersected by the
underlying surface. In contrast, NMC requires 23 edge, face, and interior vertices per grid cell~\cite[Fig.5]{chen2021nmc}.

\begin{figure}
\centering
\includegraphics[width=0.99\linewidth]{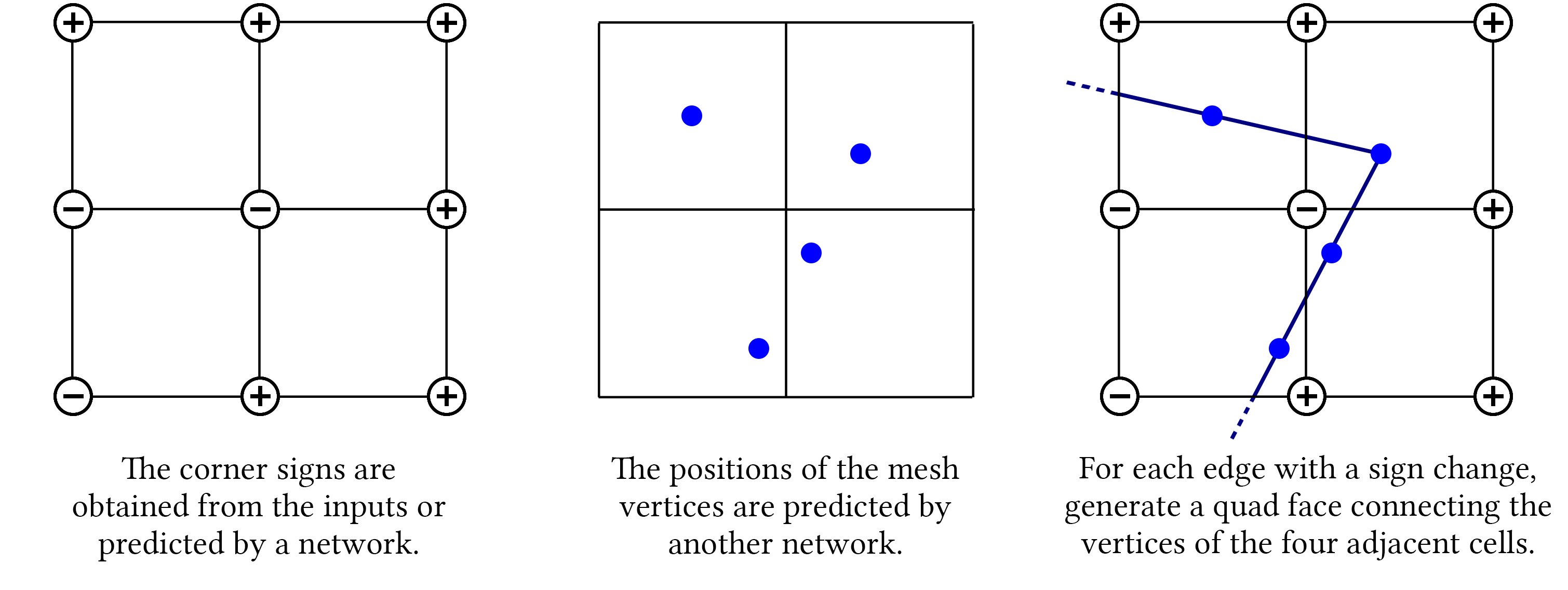}
\captionof{figure}{\textbf{Neural dual contouring~(NDC)}}
\label{fig:ndc}
\vspace{2mm}
\includegraphics[width=0.99\linewidth]{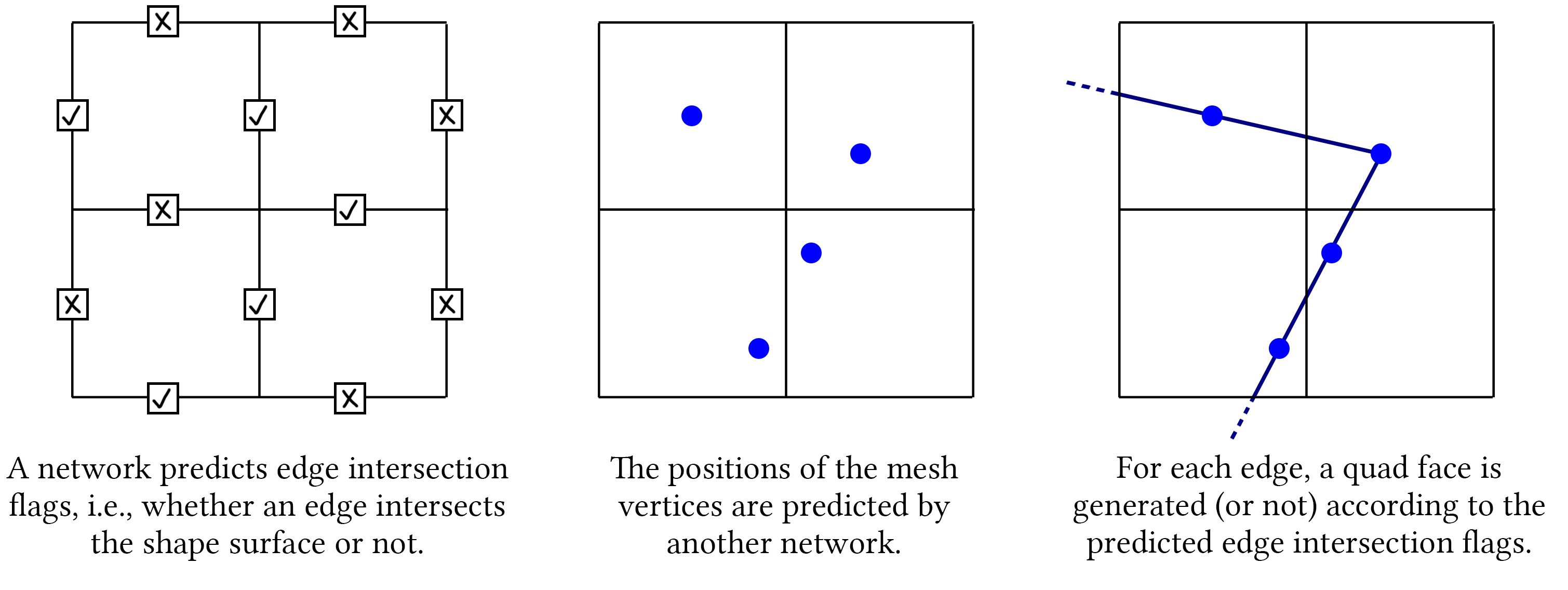}
\captionof{figure}{\textbf{Unsigned neural dual contouring (UNDC)}}
\vspace{-3mm}
\label{fig:undc}
\end{figure}

A traditional drawback of the classical DC, as compared to MC, is that it requires gradients (i.e., surface normals) as 
input to compute a suitable vertex location within each cell.  Our data-driven approach does not have this drawback.  
NDC employs a neural network trained on example 3D surface data to predict the vertex locations (\Figure{ndc}).  Our neural network
learns to compute whatever gradients and/or contexts that are useful to reproduce the training surfaces, and thus can 
operate on a voxel grid {\em without gradients as input\/}.

Another key feature of DC is that its meshing only requires knowing whether a cell edge is intersected by 
the output surface or not~\cite{li2010polygonizing}.  We can thus train our network to predict an intersection or crossing {\em flag\/} per edge, in addition 
to vertex locations, without accounting for signs at cell corners (\Figure{undc}). We refer to this version of our network as 
{\em unsigned\/} NDC, or UNDC for short.  With the {\em sign-agnostic\/} UNDC, we can forgo both the input requirement on signed 
distances and the output requirement that the resulting mesh is closed and watertight, as for MC and its variants.

Our learning model is built with 3D convolution neural networks (CNNs) separately trained for vertex prediction and the prediction of
cell corner signs (NDC) or edge crossings (UNDC)\footnote{{Note that in the rest of the paper, we use the term NDC to refer to both our overall dual contouring based learning framework {\em and\/} the
specific network that reconstructs meshes based on sign prediction (\Figure{ndc}). On the other hand, the term UNDC is used exclusively to denote the sign-agnostic
version of our method (\Figure{undc}).}}.
Our network training is supervised with an L2 reconstruction loss against
pseudo ground-truth vertices computed by DC and binary cross entropy loss for sign/crossing predictions.
As in NMC, our CNNs are designed with limited receptive fields to ensure generalizability.

We train our NDC networks on a CAD dataset, ABC~\cite{koch2019abc}, and we test them on ABC and four other datasets to assess 
generalizability: 1) Thingi10K~\cite{zhou2016thingi10k}, a dataset of 3D-printing models, 2) FAUST~\cite{FAUST}, a 
dataset of human body shapes, {3) MGN ~\cite{MGN}, a dataset of clothes as open surfaces,} and 4) Matterport3D ~\cite{Matterport3D}, 
a collection of scenes with noisy RGB-D depth images.
Quantitative and qualitative evaluations on isosurfacing using voxel data as input
suggest that NDC clearly outperforms MC33 and several variants of NMC
in terms of mesh reconstruction quality, feature preservation, triangle quality, and inference time,
when using signed (distances or binary voxels) grids as inputs. 
At the same time, NDC produces 
4-8 times fewer mesh elements using 3-20 times less inference time, compared to NMC.
Further experiments with point cloud inputs suggest that UNDC outperforms both classical non-learning
based methods, such as Ball Pivoting~\cite{bernardini1999ball},  Screened Poisson reconstruction~\cite{kazhdan2013screened}, and recent 
reconstructive neural networks such as SIREN~\cite{sitzmann2020implicit}, Local Implicit Grids~\cite{jiang2020local}, and
Convolutional Occupancy Networks~\cite{peng2020convolutional}.  Qualitative and quantitative results show significant improvements for NDC in terms of reconstruction quality, feature preservation, and inference time.
Our main contributions can be summarized as follows:
\begin{itemize}[leftmargin=*]
\setlength\itemsep{0em}
\item We propose the first data-driven approach to mesh reconstruction based on dual contouring. Unlike classical DC, which optimizes vertex locations within the confines of individual cells using a handcrafted Quadratic Error Function~(QEF)~\cite{qem}, NDC predicts vertex locations using a learned function, which eliminates the need for gradients in the input and accounts for local {\em contextual\/} information inherent in the training data.
\item A unified learning model that is applicable to a larger variety of inputs than previous meshing methods. As shown in
\Figure{teaser}, the allowed inputs include signed/unsigned distance fields, binary voxels, and un-oriented point clouds.
\item A significant, 23:1, reduction in representational complexity by NDC over NMC translates to across-the-board gains, 
in terms of simplicity of the network architecture, as well as reduction in network capacity, training and inference times, and 
more; see Table~\ref{tab:compare_nmc} for a summary.
\item A sign-agnostic network, UNDC, that can produce {\em open\/}, even {\em non-orientable\/}, output surfaces; see Figure~\ref{fig:teaser}.
\end{itemize}

\begin{table}[t]
\caption{Comparing various aspects of NMC vs.~NDC.}
\label{tab:compare_nmc}
\begin{center}
\resizebox{1.0\linewidth}{!}{
\begin{tabular}{lcc}
 & NMC & NDC \\
\midrule
Output & 5 (bool) \!+\! 51 (float) per cube & 1 (bool) \!+\! 3 (float) per cube \\
\midrule
Network & 3D ResNet & {6-layer 3D CNN} \\ 
\midrule
\multirow{2}{*}{Tessellation} & Manually designed, & {$\le 1$ vertex per cell; $\le 1$ quad} \\
 & 37 unique cases per cube & {per edge; see Figure~\ref{fig:ndc}} \\
\midrule
{Output} vertex count & $\approx 8 \times$ MC & $\approx$ MC \\
\midrule
{Output} triangle count & $\approx 8 \times$ MC & $\approx$ MC \\
\midrule
\multirow{5}{*}{Data preparation} & Sample dense point cloud & Sample only vertex \\
 & in each cube; minimize & signs, intersection points \\
 & chamfer distance via back & and normals; then apply \\
 & propagation; complex & Dual Contouring; Fast \\
 & and time-consuming & and easy to compute. \\
\midrule
\multirow{3}{*}{Implementation} & Need to consider all & Could be a nice \\
 & cube tessellation cases; & undergraduate \\
 & difficult to implement & assignment \\
\midrule
\multirow{3}{*}{Regularization} & Need a complex & No regularization  \\
 & regularization term & term needed \\
 & for voxel input & \\
\midrule
\multirow{2}{*}{Trainging time} & (On ABC training set) & (Same setting) \\
 & 4 days per network & < 12 hours per network \\
\midrule
\multirow{2}{*}{Inference speed} & ($64^3$ SDF input) & (Same setting) \\
 & > 1 second per shape & 30+ shapes per second \\
\midrule
\multirow{2}{*}{Inherent issues} & Self-intersections, thin & Non-manifold \\
 & triangles with small angles & edges and vertices \\
\bottomrule
\end{tabular}
}
\end{center}
\centering
\end{table}

\section{Related work}
\label{sec:related}

The literature on mesh reconstruction is extensive and so we refer to several surveys for full coverage~\cite{araujo2015survey,berger2017survey}.
In this section, we focus on techniques for isosurfacing (i.e., mesh extraction from discrete volume data) and surface reconstruction from point 
cloud data, with a focus on the recent data-driven approaches most closely related to our work. Then in \Section{method_dc}, we formally define 
dual contouring (DC), establish notations used throughout the paper, and compare DC to marching cubes.

\subsection{Isosurfacing and differentiable reconstruction}

The marching cubes (MC) approach for isosurfacing from discrete signed distances was first proposed concurrently by~\cite{lorensen1987marching}
and~\cite{DataStructureforSoftObjects}. Since then, many variants have followed, including the best-known
MC33~\cite{chernyaev1995marching}, which correctly enumerated all possible topological cases for mesh tessellations, based on the
trilinear interpolation assumption. Indeed, most of the MC follow-ups made the same assumption and are unable to recover sharp features.
This issue was resolved by neural marching cubes (NMC)~\cite{chen2021nmc}, which combines deep learning with MC for the first time,
building on the premise that feature recovery can be learned from training meshes.

Our work is inspired by NMC.
In NDC, we combine deep learning with dual contouring (DC)~\cite{ju2002dual} to bring key advantages of
classical DC over MC into a learned mesh reconstruction model, without requiring any additional inputs (e.g. gradients).
In addition to improved efficiency and reconstruction quality (see \Section{results}), our method also represents the first unified mesh reconstruction framework that can take on all the input types shown in \Figure{teaser}. To the best of our knowledge, no previous methods were designed to reconstruct meshes from {\em unsigned\/} distance fields.

Several recent works, including deep marching cubes (DMC)~\cite{DeepMarchingCubes}, MeshSDF~\cite{MeshSDF},
and Deformable Tetrahedral Meshes (DefTet)~\cite{gao2020deftet}, propose {\em differentiable\/} mesh reconstruction schemes. While both these
methods and NDC bring deep learning to mesh reconstruction, their focuses and strengths are quite different. DMC, MeshSDF, and DefTet
all target end-to-end differentiability, while offering limited capabilities to reconstruct geometric and topological details. They also encode \textit{global} features for their predictions, which can hinder both scalability, reconstruction quality (as downsampling is necessary during training), and generalizability. In contrast, our work focuses on learning a refined meshing model applicable to a variety of inputs. We target fine-grained
quality criteria related to feature preservation and surface quality. Our learning model is also local, 
hence highly scalable and generalizable to diverse shape types and classes.

\subsection{Mesh reconstruction from point clouds}

Many methods have been proposed for surface mesh estimation from unorganized points.   Following the taxonomy in \cite{berger2017survey}, previous works can be characterized based on the underlying priors, e.g., smoothness \cite{kazhdan2013screened},  visibility \cite{curlesslevoy}, dense sampling~\cite{AmentaVoronoi},
primitives \cite{schnabel2009completion}, and learning from data \cite{williams2019deep}.
Among the methods based on data priors, 
some compose surfaces explicitly from patches extracted from examples \cite{funkhouser2004modeling, pauly2005example, shen2012structure}.  Others learn implicit priors, either for entire objects \cite{park2019deepsdf, mescheder2019occupancy, chen2019learning, peng2021shape, chibane2020neural} or for patches \cite{badki2020meshlet, groueix2018papier, jiang2020local, mi2020cvpr, peng2020convolutional, sitzmann2020implicit, williams2019deep, Hanocka2020p2m}.   Both NMC~\cite{chen2021nmc} and NDC are in the latter category: they learn implicit priors for local regions.

Surface reconstruction methods also differ in whether they can work for input point clouds without normals \cite{atzmon2020sal, tang2021sa}, whether the output mesh interpolates the input points via triangulation~\cite{rako2021cvpr,sharp2020eccv,liu2020eccv}, and whether they can produce open surfaces from partial scans, e.g., via an advancing front scheme~\cite{cohensteiner2004,bernardini1999ball}.   Of course, normals can be estimated in a preprocessing step (e.g., using \cite{boulch2012fast}), and open surfaces can be created from watertight reconstructions in a postprocessing step (e.g., using SurfaceTrimmer in \cite{kazhdan2013screened}).  However, these separate steps rely on heuristic algorithms with parameters that are difficult to tune (e.g., size of neighborhood for normal estimation, density of points for surface trimming, etc.).  By comparison, our UNDC includes all these steps in a single learned process that can produce open, even non-orientable, meshes directly from unoriented point cloud inputs, with fast inference. Also, our method is non-interpolatory, hence insensitive to sampling non-uniformity and noise (with noise augmentation in training).
In \Section{results_pc}, we compare UNDC with several representative learning-based reconstruction networks~\cite{sitzmann2020implicit,peng2020convolutional,jiang2020local} whose results are most competitive to ours. 
Technical details about these works are described in the \SupplementaryMaterial.

\subsection{Dual Contouring~(DC)}
\label{sec:method_dc}
\citeN{ju2002dual} introduced \textit{Dual Contouring} to convert a Signed Distance Field~$\SDF:\real^3 \rightarrow \real$ into a polygonal mesh $\mathcal{M}=(\Vertices,\Faces)$; see~\Figure{mc_vs_dc} (top).
This is achieved by discretizing the function on a lattice $\Grid=(\GridVertices, \GridEdges)$.  It first samples the $\SDF$ at the grid vertices $\GridVertices$ and determines their signs $\Signs$.  Then, it finds the zero crossings $\EdgeVertices$ of the $\SDF$ on the lattice edges spanning vertices with opposite signs.   Next, it computes the gradients of the $\SDF$ at those crossings, which provide surface normals $\EdgeNormals$.  Finally, it creates quadrilateral polygonal faces $\Faces$ that are \textit{dual} to the lattice edge crossings~$\GridEdges$.
In what follows we have $|\GridVertices|=M \times N \times K$ lattice vertices, $|\GridEdges|=(M-1) \times (N-1) \times (K-1) \times 3$ lattice edges, and we index $\GridVertices$ by $(m,n,k)$, while we refer to edges as $(i,j) \in \GridEdges$.
Dual contouring assumes as input:
\begin{align}
\Signs &\in \bool^{|\GridVertices|}, & \Signs &= \fun_\Signs(\SDF, \Grid), &\text{(grid signs)}
\label{eq:dcsigns}
\\
\EdgeVertices &\in \real^{|\GridEdges| \times 3}, & \EdgeVertices &= \fun_\EdgeVertices(\SDF, \Grid), &\text{(edge vertices)}
\label{eq:dcedgevertices}
\\
\EdgeNormals &\in \real^{|\GridEdges| \times 3}, & \EdgeNormals &= \fun_\EdgeNormals(\SDF, \Grid), &\text{(edge normals)}
\label{eq:dcedgenormals}
\end{align}
where, analogously to marching cubes~\cite{lorensen1987marching}, 
$\Signs$ are the signs of~$\SDF$ on the lattice vertices, that is $\fun_\Signs:\text{sign}(\SDF(\GridVertices))$, $\fun_\EdgeVertices$ computes the zero-crossings of~$\SDF$ along the lattice edges, and $\fun_\EdgeNormals:\nabla\SDF(\EdgeVertices)$ are the gradients of~$\SDF$ measured at~$\EdgeVertices$.
Given these quantities, dual contouring generates a polygonal mesh, consisting of \textit{quad} faces and corresponding vertices:
\begin{align}
\Faces &\in \bool^{|\GridEdges|}, &\Faces &= \fun_\Faces(\Signs),
\label{eq:dcfaces}
\\
\Vertices &\in \real^{|\GridVertices| \times 3}, &\Vertices &= \fun_\Vertices(\EdgeVertices, \EdgeNormals),
\label{eq:dcvertices}
\end{align}
where, with a slight abuse of notations, we use the same nomenclature $\fun_\Faces$ for a polygonal face (i.e. tuple of vertex indices) and the Boolean value that determines whether the face should be created.

Dual faces $\Faces$ are created \textit{only} whenever lattice edges connect lattice vertices of \textit{opposite} signs $\Signs$:
\begin{align}
\fun_\Faces &: \text{xor}(\Signs_i, \Signs_j), \quad (i,j) \in \GridEdges.
\end{align}
Vertices are created by triangulating, a-la~\citeN{qem}, the planar constraints defined on the edges of each voxel in the lattice:
\begin{align}
\fun_\Vertices &: \argmin_{\mathbf{x}} \!\!\!\sum_{\edge \in \Grid_{mnk}} \!\!\! (\EdgeNormals_\edge \cdot (\mathbf{x} - \EdgeVertices_\edge))^2,
\label{eq:QEFs}
\end{align}
where $\Grid_{mnk}$ refers to the voxel rooted at $\GridVertices_{mnk}$, and $\mathbf{e}$ iterates the $12$ edges of the voxel.

\paragraph{Comparison to MC}
DC has the drawback that it assumes the availability of the function's gradients $\EdgeNormals$.
This perhaps justifies why it has not been as popular as MC, which \textit{only} requires signs \eq{dcsigns} and zero-crossings~\eq{dcedgevertices}.
Nonetheless, the mesh creation mechanism of dual contouring is \textit{significantly} simpler than the one in MC, where the former involves simple Boolean operations, while the latter involves enumerating all possible combinations and employing look-up tables that define the corresponding topology.
Further, note that MC tends to discard high frequency information (i.e. sharp corners), DC is capable of preserving such details to a much better extent.

\section{Method}
\label{sec:method}

In this paper, we introduce a learning framework, neural dual contouring (NDC), that achieves the simplicity and sharp features of DC without requiring function gradients in the input.
Given any common input representation $\Input$ (e.g. point cloud, signed or unsigned distance functions, or voxelized grids),
NDC can be formalized by a simple generalization of Equations~(\ref{eq:dcsigns}, \ref{eq:dcfaces}, \ref{eq:dcvertices}).
In particular, we introduce \textit{two} different techniques, illustrated with a 2D example in~\Figure{ndc} and~\Figure{undc}, and detailed in what follows.

The first, and default, variant of our method, which reconstructs meshes based on sign prediction, is simply referred to as NDC.
It can be algebraically formalized as:
\begin{align}
\text{NDC}(\Input) = 
\begin{cases}
\Signs \!\!\!\!&= \fun_\Signs(\Input, \Grid \given \pars),
\\
\Vertices \!\!\!\!&= \fun_\Vertices(\Input, \Grid \given \pars),
\\
\Faces \!\!\!\!&= \text{xor}(\Signs_i, \Signs_j).
\end{cases}
\label{eq:ndc}
\end{align}
The logic controlling whether a face should be generated is identical to classical DC, while vertices and signs are predicted by neural networks (with trainable parameters~$\pars$) that receive as input~$\Input$.
At the same time, the input requirements of \NDC are closer to the ones of MC and NMC: we do not require the availability of normals as in classical DC since we do not perform explicit optimizations for vertex locations using~\eq{QEFs}.  Instead, vertex positions are predicted with a network trained from examples.

The second variant, named \UNDC, with U denoting ``unsigned'', is similar to NDC, but it \textit{directly} predicts the existence of dual faces $\Faces$ rather than resorting to sign prediction:
\begin{align}
\text{UNDC}(\Input) = 
\begin{cases}
\Vertices \!\!\!\!&= \fun_\Vertices(\Input, \Grid \given \pars),
\\
\Faces \!\!\!\!&= \fun_\Faces(\Input, \Grid \given \pars).
\end{cases}
\label{eq:undc}
\end{align}
The key advantage of this variant is that it can produce surface crossings without having to rely upon differences of inside/outside signs at grid cell vertices.   This feature allows UNDC to operate on \textit{unsigned} distance fields or \textit{non-oriented} point clouds (we employ the prefix \textit{U} to indicate this variant's ability to operate on \textit{unsigned} inputs).  It also allows UNDC to produce mesh faces, likely in the form of {\em thin sheets\/}, in regions where the underlying object parts are {\em thinner\/} than one voxel. Clearly, such thin parts are not representable by differences of grid vertex signs, and as a result, methods including MC, NMC, as well as NDC, would not be able to reconstruct them at all; see \Figure{result_sdf} in \Section{results}.  Additionally, UNDC can produce open surfaces with boundaries directly for input data representing partial surfaces.   These advantages are in contrast to other methods like MC and variants that guarantee their outputs to be watertight and can represent only solid objects without thin features.

\subsection{Encoders}
Let us now consider the design of $\fun_\Vertices$, $\fun_\Signs$, and $\fun_\Faces$ for different types of input $\Input$:
\CIRCLE{1} signed/unsigned distance functions,
\CIRCLE{2} voxelized occupancy, and
\CIRCLE{3} point clouds.

\begin{figure}
\centering
\includegraphics[width=0.99\linewidth]{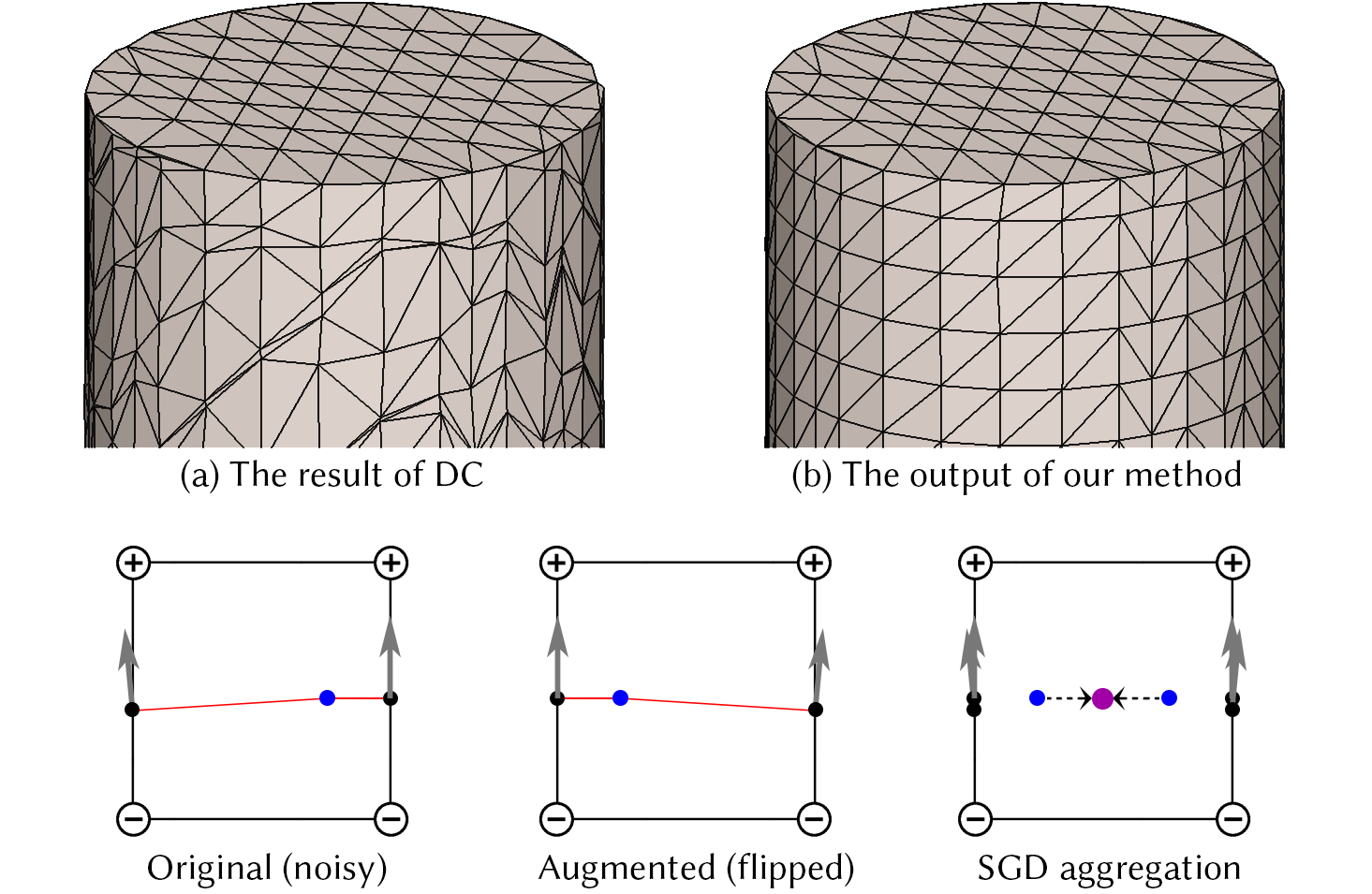}
\caption{
\textbf{Training data preparation with data augmentation} -- 
The ground truth meshes computed using classical DC (a) can be noisy. With proper augmentation for the training data (see bottom), our NDC network can be trained to output meshes with better tessellation quality (b).}
\label{fig:noisydc}
\end{figure}

\paragraph{Distance Function Inputs}
When a Signed Distance Function (SDF) $\SDF$ is provided as input, our model $\fun_\Vertices$ first samples the function $\SDF$ at the grid vertices $\GridVertices$ into a floating point tensor of shape $|\GridVertices|$.
We then use a 3D CNN to process this tensor; the 3D CNN has 6 layers, with the first 3 layers having kernel size $3^3$ and the last 3 layers having kernel size $1^3$, an overall receptive field of $7^3$.
We employ hidden layers with $64$ channels to make the network computationally efficient (i.e. 37 fps) as it has few network weights (i.e. 1MB).
Leaky ReLU activation functions are employed everywhere except at the output layer where sigmoids are used.
Note that when \NDC operates on SDFs, $\fun_\Signs$ is extremely efficient as it just requires the computation of a sign at lattice locations similarly to classical dual contouring.
Finally, the architecture of $\fun_\Faces$ for the \UNDC model is the same as $\fun_\Vertices$ in the \NDC model.

\paragraph{Voxelized Occupancy Inputs}
For this class of inputs, we use a network with almost the same architecture as for SDF input, but with a small modification to enlarge the receptive field to $15^3$ (i.e. employ 7 rather than three $3^3$ convolutional layers).
Our rationale is that voxelized occupancies are heavily quantized, and a larger receptive field would allow the network to develop stronger \textit{priors} to cope with the larger degree of ambiguity in the data.

\begin{figure*}
  \centering
  \includegraphics[width=0.99\linewidth]{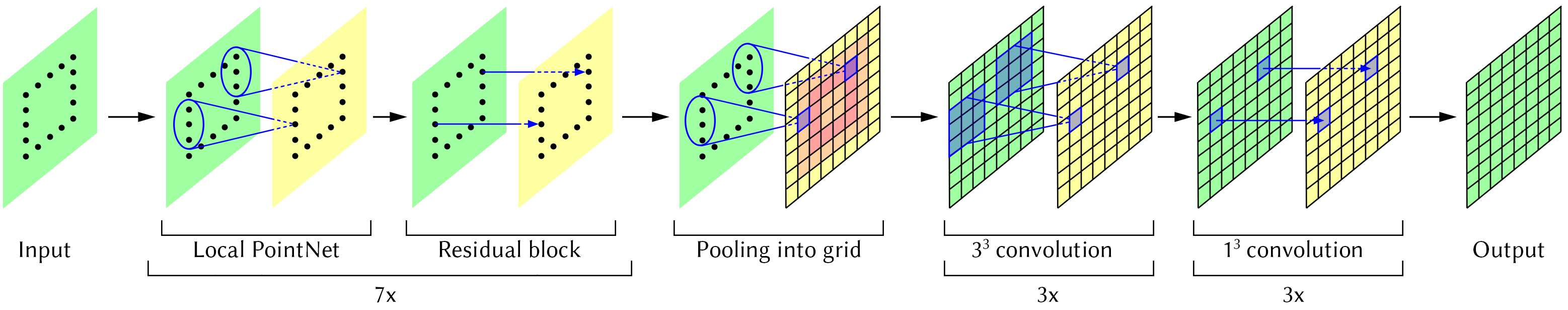}
	\caption{The architecture of our point cloud processing network for UNDC.}
	\label{fig:network}
\end{figure*}

\paragraph{Point Cloud Inputs}
For point cloud inputs, we devise a local point cloud encoder network divided into two parts: 
\CIRCLE{1} point cloud processing and \CIRCLE{2} regular grid processing.
The former is implemented as a dense PointNet++ ~\cite{pointnet++}, while grid processing has three $3^3$ convolution layers and three $1^3$ convolution layers, hence of a similar architecture to the one used for inputs represented as grids. 
The network architecture is shown in \Figure{network}. Further details are available in the \SupplementaryMaterial.

\subsection{Training data preparation}
\label{subsec:training_data}

To obtain the training data, we place random 3D mesh objects in the grid $\Grid$ to compute signs, intersection points, and corresponding ground truth normals\footnote{Note that these intersection points and normals were utilized to create the pseudo-ground truth at training time; they are not used at test time.}, and then apply classical DC to obtain the ground truth vertex predictions.
However, this process can result in aliased normals, which can lead to poorly positioned vertices after the optimization in~\eq{QEFs}; see \Figure{noisydc}-(left).

While this situation might seem problematic at first, we make the same observation made by~\citeN{lehtinen2018noise2noise} in this setting.
In particular, the use of an L2 reconstruction loss, coupled with data augmentation, leads to a zero-mean distribution in the vertex positions predicted by our model; see~\Figure{noisydc}-(b).
To achieve this zero-mean distribution of optimization residuals, we augment the training data by rotation (by $\pi/2$ around the Euclidean axes), mirroring, and (global) sign inversion.
Note this augmentation is \textit{not} done within a mini-batch, but rather, we rely on stochastic gradient descent for aggregating data towards a zero-mean residual configuration progressively over the course of training.

\subsection{Training losses}
Given ground truth data, note that all the sub-networks within the \NDC and \UNDC models can be trained \textit{separately}, leading to a simpler training setup where no hyper-parameter tuning between losses becomes necessary.
Note that we leverage our input data to only supervise the prediction made by the networks in a narrow-band around the input surface, with binary masks~$\Masks_\Signs$, $\Masks_\Vertices$ that evaluate to one if we are within the narrow-band, and zero otherwise; see the \SupplementaryMaterial for additional details.
This is because surfaces should only be created in the proximity of either changes in the sign of $\SDF$, in occupancy for voxelized inputs, or proximity of the input points for point cloud inputs.
We start with a simple L2 reconstruction loss of pseudo ground-truth vertices (i.e. as computed by dual contouring):
\begin{equation*}
\loss{\Vertices}(\pars) = \mathbb{E}_{(\Input, \Masks_\Vertices, \Vertices_\gt) \sim \mathcal{D}}
\sum_{m,n,k} \left[ \| \Masks_\Vertices \odot (\fun_\Vertices(\Input, \Grid \given \pars) - \Vertices_\gt) \|_2^2  \right],
\label{eq:MSE}
\end{equation*}
where $\odot$ is the Hadamard product on $\Grid$.
For \NDC, we supervise the prediction of signs via Binary Cross Entropy~(BCE):
\begin{equation*}
\loss{\Signs}(\pars) = \mathbb{E}_{(\Input, \Masks_\Signs, \Signs_\gt) \sim \mathcal{D}}
\sum_{m,n,k} \left[ \Masks_\Signs \odot \BCE \left(\fun_\Signs(\Input, \Grid \given \pars), \Signs_\gt \right) \right].
\label{eq:BCE}
\end{equation*}
Finally, for \UNDC the loss $\loss{\Faces}(\pars)$ is analogous to $\loss{\Signs}(\pars)$, and hence we do not repeat its definition.

\begin{figure}
\centering
\includegraphics[width=0.99\linewidth]{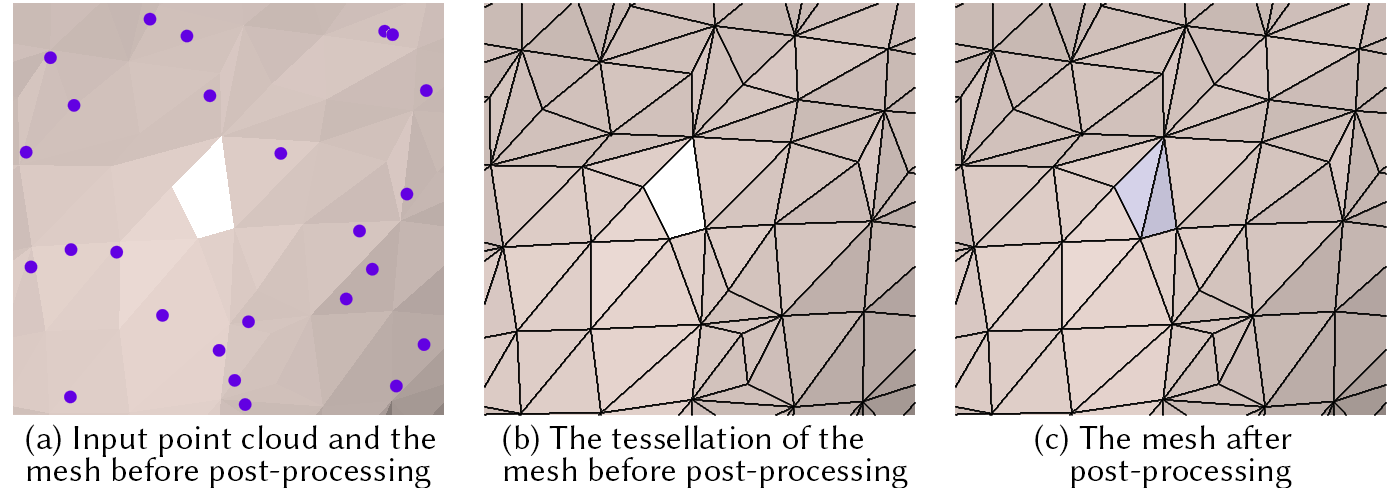}
\caption{
\textbf{Post-processing UNDC outputs} -- 
The post-processing step can close small holes by adding quad faces.}
\label{fig:postprocessing}
\end{figure}
\subsection{Post-processing} 
When \UNDC is operating on sparse or noisy point clouds, the function $\fun_\Faces(\Input, \Grid \given \pars)$ that predicts grid edge crossings can make mistakes, leading to small holes in the output mesh.
Empirically, the holes are typically small and isolated (see Table~\ref{tab:open_surface_non_manifold}), and so we can use a simple post-processing step to close them.
We employ our tensor representation of the mesh $\Faces \in \bool^{|\GridEdges|}$ to determine boundary edges from $\mathcal{M}$, and we flip Boolean entries in $\Faces$ that would result in three/four edges to change their boundary state; see \Figure{postprocessing}.
These post-processing steps are executed on the GPU, resulting in a negligible impact on the overall inference time.

\section{Results and evaluation}
\label{sec:results}

We ran a series of experiments with NDC and UNDC to evaluate their performance in comparison to previous methods for a variety of input types, including SDFs, unsigned distance fields (UDF), binary voxel grids, point clouds, and depth image scans.

\subsection{Datasets, training, and evaluation metrics}
\label{sec:results_data}

In all of our experiments, we train NDC and UNDC on the ABC dataset \cite{koch2019abc} following the protocols in NMC \cite{chen2021nmc}. The ABC dataset consists of watertight triangle meshes of CAD shapes, which are characterized by their rich geometric features including both  sharp edges and smooth curves, as well as their topological varieties. We only use the first chunk of ABC dataset for our experiments. We split the set into 80\% training (4,280 shapes) and 20\% testing (1,071 shapes). During the data preparation, we obtain meshes over $32^3$ and $64^3$ grids to train our network. We evaluate the methods on the \textit{test set} of ABC.

\paragraph{Generalization}
To assess generalization capabilities of the methods, we evaluate on four other datasets, also following the experimental settings as in NMC \cite{chen2021nmc}.
The additional test sets include 2,000 shapes from Thingi10K dataset ~\cite{zhou2016thingi10k}, a dataset of 3D-printing models; 100 shapes of human bodies from FAUST dataset~\cite{FAUST}, a dataset of organic shapes; several shapes in MGN \cite{MGN}, a dataset of clothes with open surfaces; and several rooms from Matterport3D \cite{Matterport3D}, a dataset containing scans of indoor scenes acquired with depth cameras. In all cases, we evaluate NDC and UNDC after training on the ABC training set without any fine-tuning.

\subsection{Metrics}
We evaluate surface reconstructions quantitatively by sampling 100K points uniformly distributed over the surface of the ground truth shape and the predicted shape, and then computing a suite of metrics that evaluate different aspects of the reconstruction. The metrics are divided into five groups.

\paragraph{Reconstruction accuracy}
We use Chamfer Distance (CD) and F-score (F1) to evaluate the overall quality of a reconstructed mesh. The metrics are good at capturing significant mistakes such as missing parts, but may not be informative for evaluating the visual quality. Therefore, we introduce other metrics to evaluate sharp feature preservation and surface quality.

\paragraph{Sharp feature preservation}
We follow NMC~\cite{chen2021nmc} and use Edge Chamfer Distance (ECD) and Edge F-score (EF1) to evaluate the preservation of sharp edges.
For a given shape, points are sampled near sharp edges and corners to form a set of edge samples.
The ECD and EF1 between two shapes are simply the CD and F1 between their edge samples.

\paragraph{Surface quality}
As in many other papers, we use Normal Consistency (NC) to evaluate the quality of the surface normals.
However, NC is similar to CD and F1 in that it mainly captures significant mistakes and neglects small mistakes which contribute significantly to visual artifacts.
Therefore, we break down NC to show the percentage of inaccurate normals (\% Inaccurate Normals, or \%IN) according to a threshold.
To compute \%IN, for each point sampled from shape $A$, we find its closest point in the points sampled from shape $B$, and then compute their angle.
If the angle is larger than the threshold, the point from $A$ is labeled as having an inaccurate normal.
\%IN (gt) is the percentage of points sampled from the ground truth shape that have inaccurate normals.
\%IN (pred) can be obtained similarly on points from the predicted shape.\footnote{Note that these two metrics are the surface normal counterparts of the two terms assembling the symmetric Chamfer Distance.}
Another aspect of mesh quality is the number of small angles in the reconstructed triangles. Therefore, we also report the percentage of small angles that are smaller than a threshold, as \% Small Angles, or \%SA.

\paragraph{Triangle \& vertex counts}
We count the number of vertices (\#V) and triangles (\#T) in the output shape to reveal the fidelity-complexity trade-off.
Note that while our method generates quad faces, we always randomly split each quad into two triangles for evaluations and visualizations.

\paragraph{Inference time}
We report inference times (seconds per shape) for the methods tested on the ABC test set. Timings are collected on the same machine with one NVIDIA GTX 1080ti GPU.

\subsection{Reconstruction from SDF} 
\label{sec:results_sdf}

\begin{figure}
  \centering
  \includegraphics[width=1.0\linewidth]{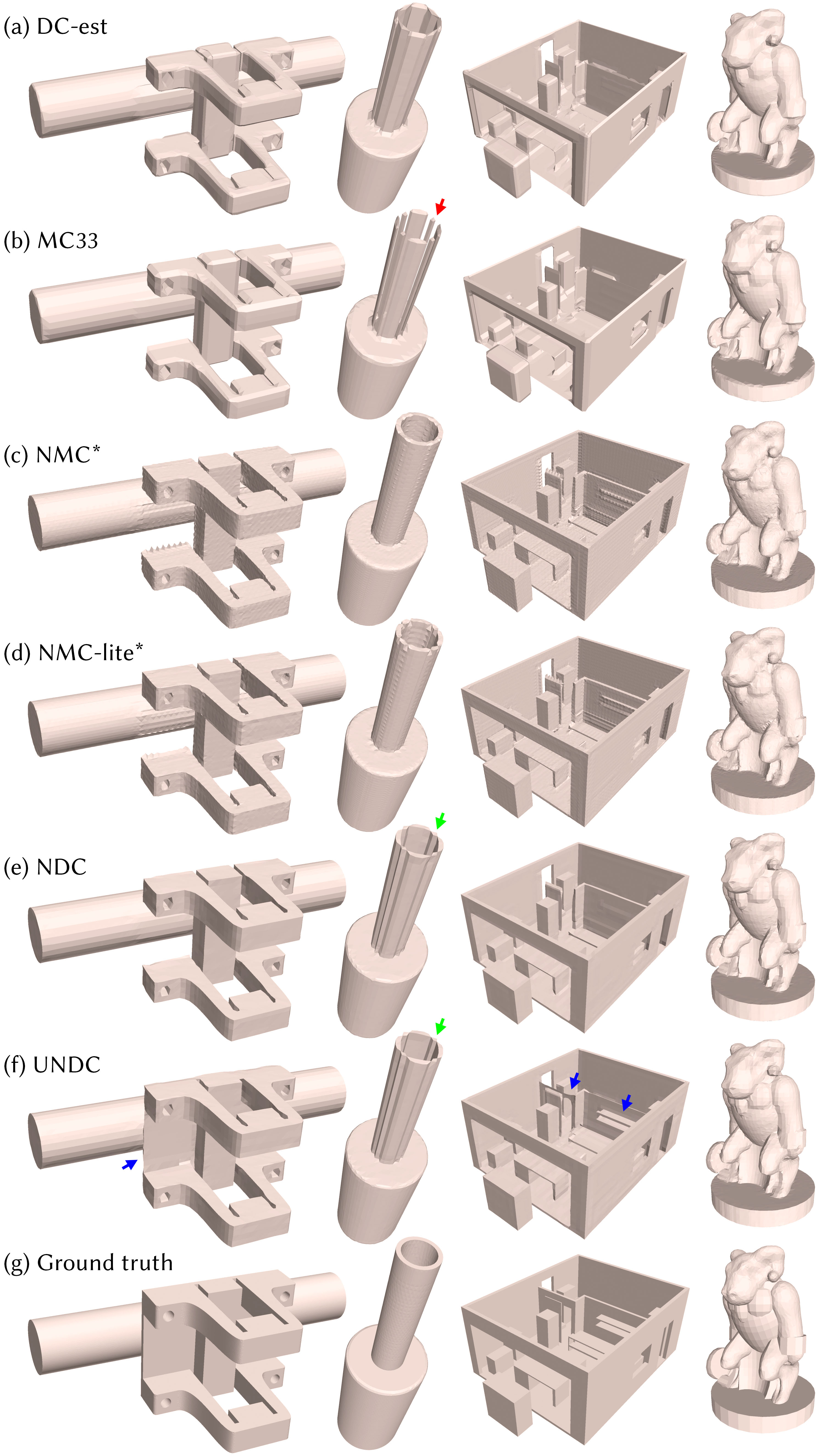}
	\caption{
	Mesh reconstruction results from \textbf{SDF} grid inputs at a relatively low resolution of $64^3$. The shapes in the first three columns are from \textbf{ABC} test set, and the last column from \textbf{Thingi10K}. Zoom in to see various surface artifacts and artifacts near edges on NMC-lite* and NMC* results, broken meshes from MC33 (red arrows), and non-manifold edges from NDC and UNDC (green arrows). Pay special attention to the thin sheets (blue arrows) reconstructed by the sign-agnostic UNDC, which correspond to parts of the ground truth shape that are thinner than one voxel. In contrast, none of the other methods (a-e) could even recover any of these thin parts.
	}
	\label{fig:result_sdf}
\end{figure}

\begin{figure}
  \centering
  \includegraphics[width=1.0\linewidth]{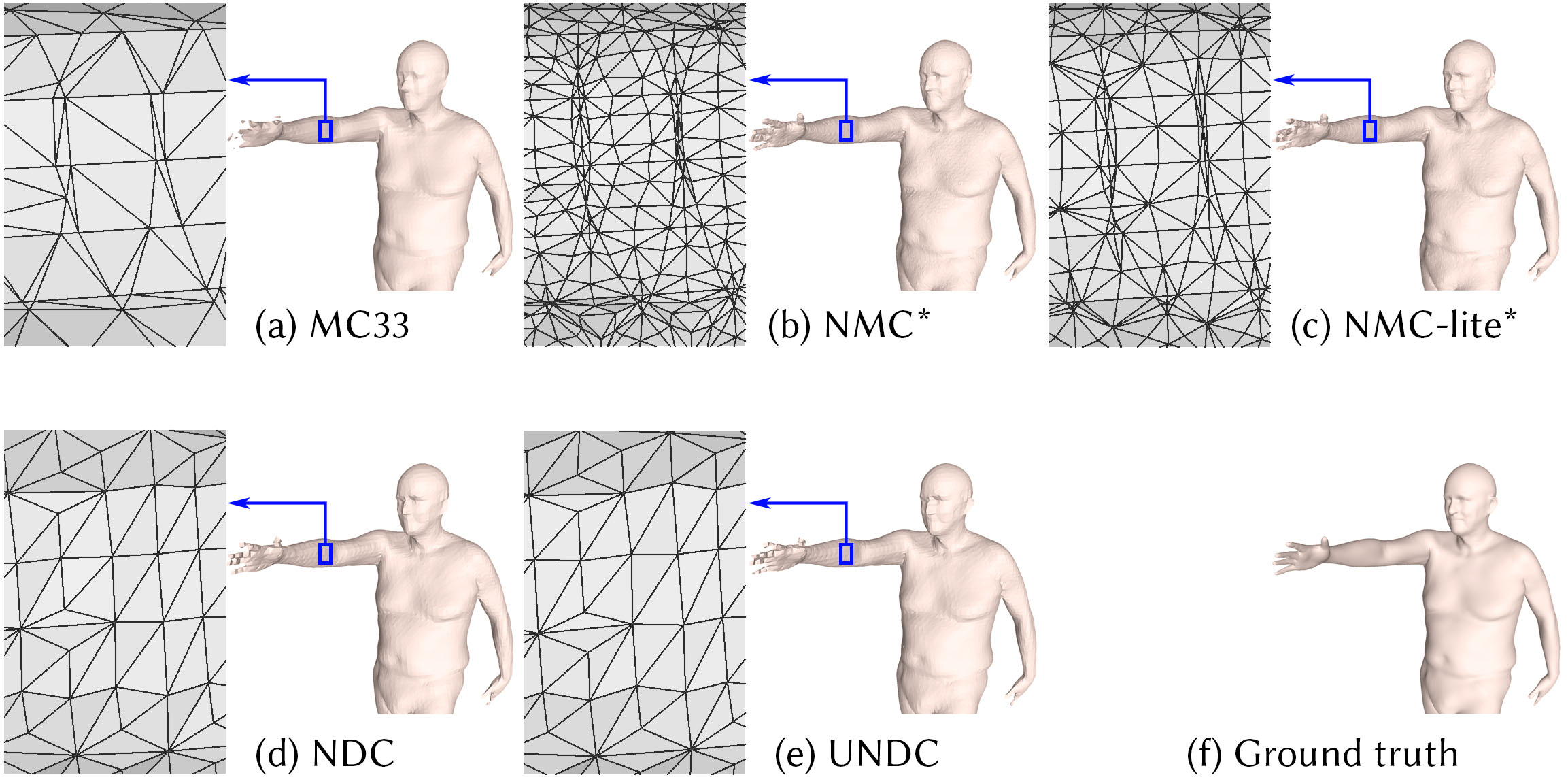}
	\caption{
	Mesh reconstruction results from \textbf{SDF} grid inputs at $128^3$ resolution on the \textbf{FAUST} dataset; see insets to compare triangle quality.
	}
	\label{fig:result_sdf_faust}
\end{figure}

\begin{table}
\captionof{table}{
Quantitative evaluation on \textbf{ABC} with \textbf{SDF} (signed or unsigned) inputs at two resolutions, evaluated on the test set split, using mesh quality metrics, output complexity, and inference times.
}
\label{tab:compare_sdf_abc}
\resizebox{1.0\linewidth}{!}{
\begin{tabular}{lrrrrrrrrrrrrrrrrr}
\toprule
$64^3$ & CD$\downarrow$ & F1$\uparrow$ & NC$\uparrow$ & ECD$\downarrow$ & EF1$\uparrow$ & \#V & \#T & {\small Inference} \\
{\small SDF input} & {\small($\times 10^5$)} & & & {\small($\times 10^2$)} & & & & {\small time} \;\; \\
\midrule
NMC         & 4.365 & 0.878 & 0.976 & 0.340 & 0.766 & 42,767 & 85,544 & 1.148s \\
NMC-lite    & 4.356 & 0.878 & 0.975 & 0.338 & 0.767 & 21,933 & 43,877 & 1.135s \\
\midrule
DC-est      & 4.673 & 0.827 & 0.958 & 3.810 & 0.167 & {\bf 5,459} & 10,969 & 0.421s \\
MC33        & 4.873 & 0.788 & 0.950 & 5.759 & 0.103 & 5,473 & {\bf 10,954} & {\bf 0.005s} \\
NMC*        & 4.400 & 0.874 & 0.972 & 0.409 & 0.715 & 42,767 & 85,544 & 0.158s \\
NMC-lite*   & 4.386 & {\bf 0.875} & 0.973 & 0.416 & 0.725 & 21,933 & 43,877 & 0.153s \\
NDC       & 4.463 & 0.867 & 0.970 & 0.338 & 0.745 & {\bf 5,459} & 10,969 & 0.027s \\
UNDC       & {\bf 0.930} & 0.873 & {\bf 0.974} & {\bf 0.328} & {\bf 0.746} & 5,584 & 11,295 & 0.051s \\
\midrule
UNDC (UDF) & 0.960 & 0.868 & 0.971 & 0.379 & 0.735 & 5,692 & 11,420 & 0.053s \\
\toprule
\toprule
$128^3$ & CD$\downarrow$ & F1$\uparrow$ & NC$\uparrow$ & ECD$\downarrow$ & EF1$\uparrow$ & \#V & \#T & {\small Inference} \\
{\small SDF input} & {\small($\times 10^5$)} & & & {\small($\times 10^2$)} & & & & {\small time} \;\; \\
\midrule
NMC         & 4.129 & 0.882 & 0.979 & 0.204 & 0.806 & 175,926 & 351,867 & 8.991s \\
NMC-lite    & 4.117 & 0.882 & 0.979 & 0.231 & 0.808 & 88,419 & 176,853 & 8.984s \\
\midrule
DC-est      & 4.132 & 0.879 & 0.977 & 2.215 & 0.266 & 22,088 & 44,213 & 1.765s \\
MC33        & 4.144 & 0.870 & 0.972 & 4.247 & 0.193 & {\bf 22,048} & {\bf 44,107} & {\bf 0.030s} \\
NMC*        & 4.116 & 0.882 & 0.978 & 0.257 & 0.779 & 175,926 & 351,867 & 1.126s \\
NMC-lite*   & 4.114 & 0.882 & 0.979 & 0.283 & 0.785 & 88,419 & 176,853 & 1.112s \\
NDC       & 4.131 & 0.881 & 0.978 & 0.214 & 0.802 & 22,088 & 44,213 & 0.207s \\
UNDC       & {\bf 0.789} & {\bf 0.890} & {\bf 0.983} & {\bf 0.149} & {\bf 0.813} & 22,578 & 45,411 & 0.410s \\
\midrule
UNDC (UDF) & 0.792 & 0.889 & 0.983 & 0.227 & 0.810 & 22,874 & 45,715 & 0.409s \\
\bottomrule
\end{tabular}
}

\vspace{4mm}

\captionof{table}{Quantitative results on \textbf{Thingi10K} with \textbf{SDF} input.}
\label{tab:compare_sdf_thing}
\resizebox{1.0\linewidth}{!}{
\begin{tabular}{lrrrrrrrrrrrrrrrrr}
\toprule
$128^3$ & CD$\downarrow$ & F1$\uparrow$ & ECD$\downarrow$ & EF1$\uparrow$ & \#V & \#T & \% IN & \% SA \\
{\small SDF input} & {\small($\times 10^5$)} & & {\small($\times 10^2$)} & & & & {\small$>5^{\circ}$} & {\small$<10^{\circ}$} \\
\midrule
MC33        & 2.421 & 0.890 & 2.657 & 0.197 & 22,324 & 44,656 & 19.08 & 2.43 \\
NMC*        & 2.613 & 0.902 & 0.269 & 0.760 & 169,211 & 338,427 & 20.99 & 0.77 \\
NMC-lite*   & 2.651 & 0.902 & 0.254 & 0.772 & 89,260 & 178,527 & 17.04 & 1.74 \\
NDC       & 2.300 & 0.901 & 0.215 & 0.792 & {\bf 22,295} & {\bf 44,631} & {\bf 12.52} & {\bf 0.24} \\
UNDC       & {\bf 0.757} & {\bf 0.904} & {\bf 0.189} & {\bf 0.795} & 22,478 & 45,043 & 12.66 & 0.29 \\
\midrule
UNDC (UDF) & 0.748 & 0.903 & 0.222 & 0.785 & 22,784 & 45,395 & 13.19 & 0.28 \\
\bottomrule
\end{tabular}
}

\vspace{4mm}

\captionof{table}{Quantitative results on \textbf{FAUST} with \textbf{SDF} input.}
\label{tab:compare_sdf_faust}
\resizebox{1.0\linewidth}{!}{
\begin{tabular}{lrrrrrrrrrrrrrrrrr}
\toprule
$128^3$ & CD$\downarrow$ & F1$\uparrow$ & ECD$\downarrow$ & EF1$\uparrow$ & \#V & \#T & \% IN & \% SA \\
{\small SDF input} & {\small($\times 10^5$)} & & {\small($\times 10^2$)} & & & & {\small$>5^{\circ}$} & {\small$<10^{\circ}$} \\
\midrule
MC33        & 0.453 & 0.985 & 0.086 & 0.387 & 12,551 & {\bf 25,076} & {\bf 34.28} & 4.23 \\
NMC*        & 0.385 & 0.990 & 0.146 & 0.552 & 83,024 & 166,038 & 44.58 & 1.18 \\
NMC-lite*   & 0.381 & 0.991 & 0.119 & 0.567 & 50,207 & 100,404 & 38.33 & 2.63 \\
NDC       & 0.397 & 0.989 & 0.044 & 0.530 & {\bf 12,538} & 25,100 & 38.38 & {\bf 0.11} \\
UNDC       & {\bf 0.362} & {\bf 0.992} & {\bf 0.038} & {\bf 0.574} & 12,609 & 25,258 & 37.38 & 0.16 \\
\midrule
UNDC (UDF) & 0.365 & 0.991 & 0.045 & 0.549 & 12,682 & 25,293 & 38.72 & 0.21 \\
\bottomrule
\end{tabular}
}

\end{table}

\begin{figure}
  \centering
  \includegraphics[width=1.0\linewidth]{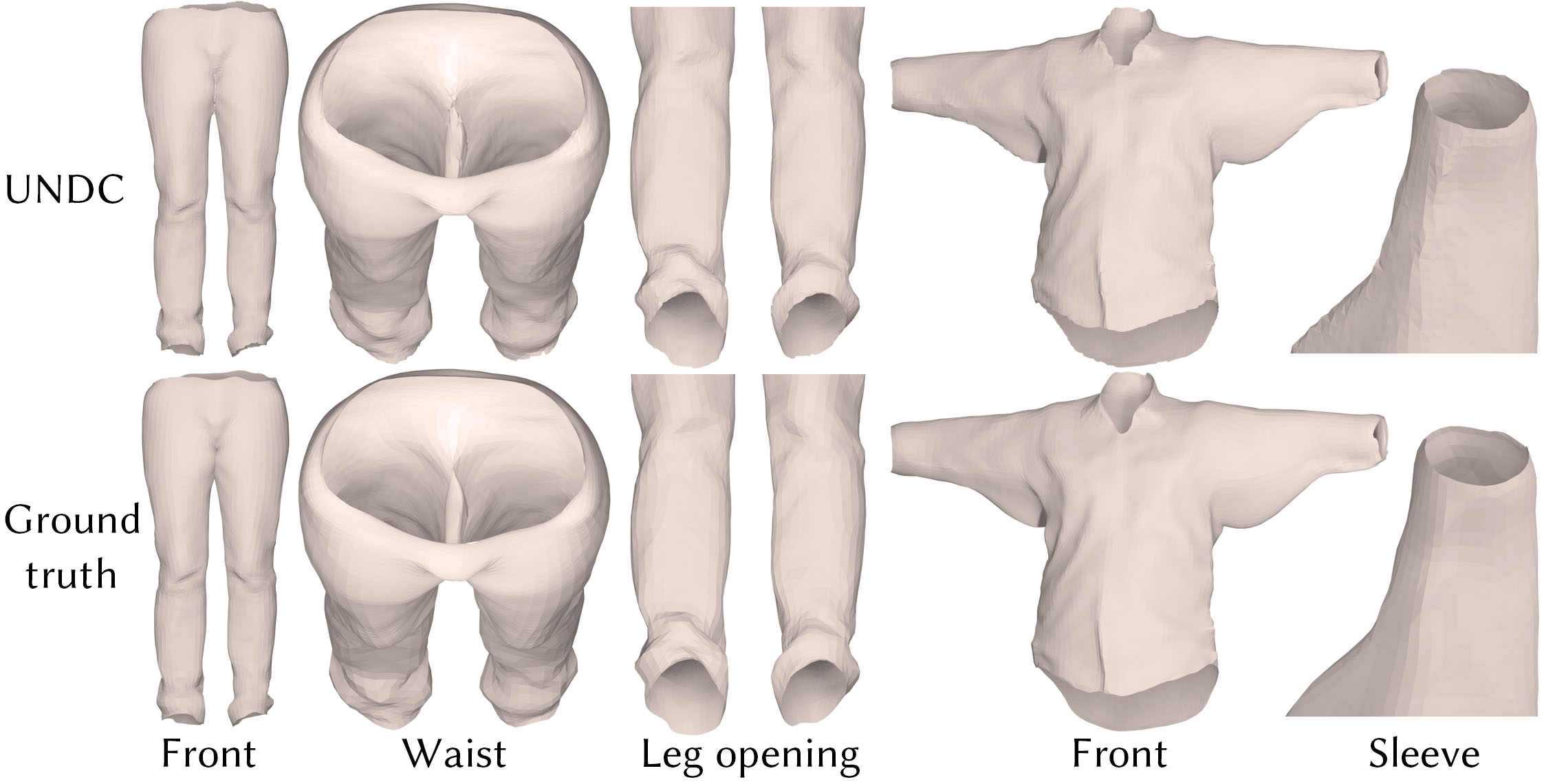}
	\caption{Qualitative results of mesh reconstruction from \textbf{UDF} inputs at $128^3$ resolution on two cloth shapes from the \textbf{MGN} dataset. Note the open surfaces reconstructed by our sign-agnostic method \textbf{UNDC}.}
	\label{fig:udf_clothes}
\end{figure}
\begin{figure}
  \centering
  \includegraphics[width=1.0\linewidth]{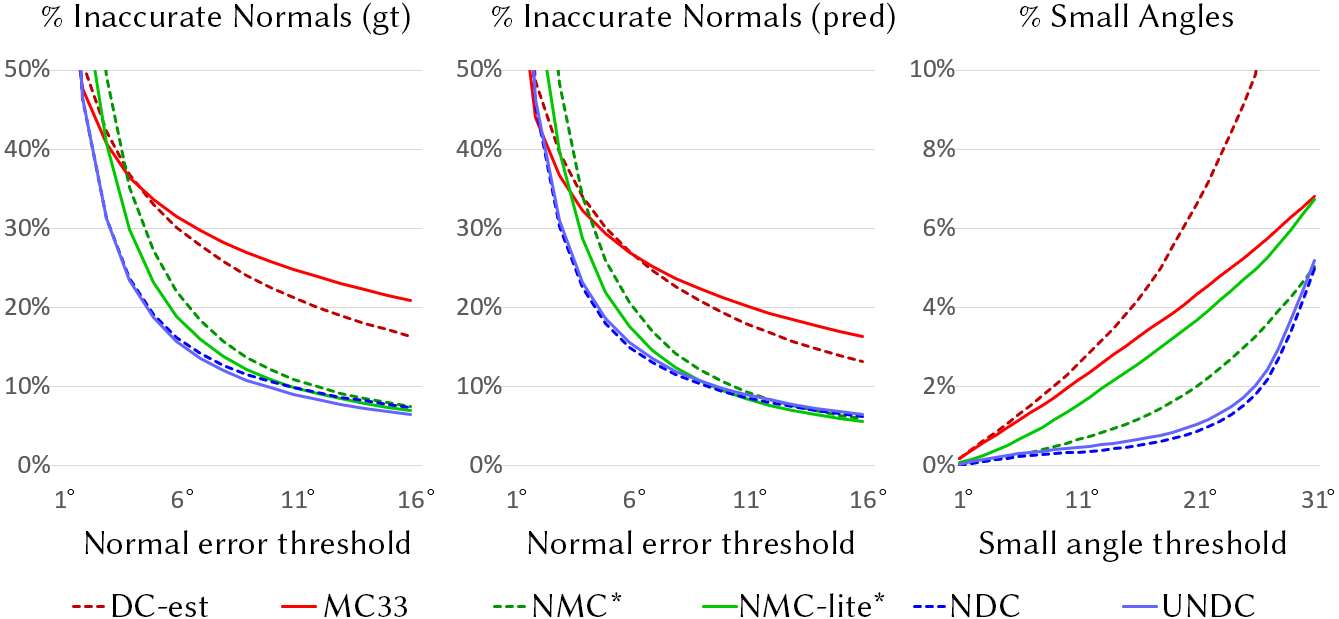}
	\caption{Some plots of surface quality (via \% of Inaccurate Normals) and triangle quality (via \% of Small Angles), on \textbf{ABC} test set with $64^3$ \textbf{SDF} input. NDC and UNDC consistently outperform other isosurfacing methods.}
	\label{fig:normal_angle}
\end{figure}

We first test NDC and UNDC on mesh reconstruction from grids of signed distances, and compare them to Marching Cubes 33 (\textbf{MC33}) (an improved version of MC to guarantee topological correctness in each cube~\cite{chernyaev1995marching,lewiner2003efficient}), classical DC with estimated normals (\textbf{DC-est}), and two versions of Neural Marching Cubes (\textbf{NMC} and \textbf{NMC-lite}) ~\cite{chen2021nmc}. 

DC-est takes the same SDF input as our method, obtains gradient values at grid points by local differentiation over the SDF, and runs the classical DC~\cite{ju2002dual} as described in Figure~\ref{fig:mc_vs_dc} by
estimating the intersection points and their gradients via linear interpolation.
Since NMC and NMC-lite use large networks to ensure the quality of the output meshes, which makes the inference significantly slower, we replace the large backbone networks in them with our 6-layer CNN to obtain \textbf{NMC*} and \textbf{NMC-lite*}, in order to have a fair comparison on reconstruction quality with respect to the inference time.
In addition, we include the results of UNDC when the input is a grid of unsigned distances as \textbf{UNDC (UDF)}. Since there is no method to reconstruct meshes from grids of unsigned distances to our knowledge, we do not compare it with other methods.

We show visual results in Figure~\ref{fig:result_sdf}, with more results provided in the \SupplementaryMaterial. We report the quantitative results on the ABC test set in Table~\ref{tab:compare_sdf_abc}. To make the paper compact, we reduce the sizes of the tables by removing some less representative metrics. The full tables can be found in the \SupplementaryMaterial.

\paragraph{Reconstruction accuracy}
NDC and UNDC consistently outperform model-driven MC33 and DC-est in terms of CD and F1.
Clearly, normal estimation is not expected to be accurate, especially near sharp features. The results of DC-est are similar to or slightly better than those of MC33, as shown in Figure~\ref{fig:result_sdf} and Table~\ref{tab:compare_sdf_abc}.

Although the network size has been significantly reduced in NMC* and NMC-lite*, they usually have slightly better results than NDC, since, given the same input resolution, Marching Cubes methods are able to reconstruct more inner structures inside each cube with their abundant tessellation templates, while NDC cannot due to its simple tessellation design. However, NMC and NMC-lite are significantly worse than UNDC in CD, even with their original large networks, due to the fact that UNDC can reconstruct thin structures which NMC methods and NDC cannot.
Visual results in Figure~\ref{fig:result_sdf} show some examples.
Note in the first and the third columns (blue arrows), some thin structures are not reconstructed by any methods other than UNDC. In the fourth column and in Figure~\ref{fig:result_sdf_faust} , we show that even on smooth shapes, NDC and UNDC can preserve more details, such as the crevices, compared to MC33. However, in the second column (green arrows), we show failure cases from NDC and UNDC, where the walls of the tube are merged together with non-manifold edges, as a result of their simpler tessellations in each cube -- this problem does not occur in NMC*. 

\paragraph{Sharp features}
As shown in Tables~\ref{tab:compare_sdf_abc},~\ref{tab:compare_sdf_thing}, and~\ref{tab:compare_sdf_faust},
NDC and UNDC consistently outperform MC33, DC-est, NMC*, and NMC-lite*, in ECD and EF1, for feature preservation. The only exception is the EF1 in Table~\ref{tab:compare_sdf_faust}, which may be due to NDC's tessellation being too simple to handle the fine structures of human shapes, e.g., the fingers.

\paragraph{Normal quality}
Since the compared methods generally have similar point-wise reconstruction accuracies, the quality of the generated surfaces in terms of visual appearance can be a better differentiator. We consistently observe that after switching to smaller networks, NMC* and NMC-lite* tend to generate noisy surfaces even over flat regions, as shown in Figure~\ref{fig:result_sdf} (c-d).
We use \%IN (IN = inaccurate normals) as a means to quantify the quality of surface normals, which correlate with surface quality. Figure~\ref{fig:normal_angle} shows the \%IN-threshold curves on the ABC test set, where the normal errors of NDC and UNDC are noticeably less than those from other methods for small threshold values, which is consistent with our visual observation that NMC* and NMC-lite* outputs exhibit more surface artifacts.

\paragraph{Triangle quality}
In Figure~\ref{fig:normal_angle}, we show \%SA-threshold (SA = small angles) curves for various methods on the ABC test set, showing that NDC outperforms the others in triangle quality, with UNDC coming close. A visual comparison can be found in Figure~\ref{fig:result_sdf_faust}.

\paragraph{Triangle and vertex count}
The \#V and \#T in Table ~\ref{tab:compare_sdf_abc} show that the vertex and triangle numbers of NDC and UNDC are very similar to those of MC33. NMC and NMC-lite produce more vertices and triangles due to their complex cube tessellation templates.

\paragraph{Inference time}
With no deep networks involved, MC33 is undoubtedly the fastest, as shown by the inference times in Table ~\ref{tab:compare_sdf_abc}. With our light network design, NDC and UNDC are next in line in terms of speed. Since NDC does not need sign predictions, it is half the size of and twice as fast as UNDC, running in real time on an NVIDIA GTX 1080ti GPU. With the newer RTX 3090 being twice as fast as GTX 1080ti, we expect UNDC to also run in real time on a higher-end GPU. In comparison, the original NMC and NMC-lite require more than a second to test on an input grid of $64^3$. Even after we replace their networks with our light designs, NMC* and NMC-lite* are still 2-4 times slower than UNDC and NDC, due to their more complex cube tessellations.
Finally, the inference time for classical DC is far from optimal as we employed our own implementation of DC-est with an unoptimized QEF solver.

\paragraph{Robustness to translation and rotation}
We highly encourage the readers to watch the video in the \SupplementaryMaterial where we test the methods on a shape while moving and rotating the shape inside the sampling grid. It clearly shows that NDC and UNDC are the most robust compared to others.

\paragraph{Varying input grid resolutions}
We report quantitative results obtained by the various methods on $64^3$ and $128^3$ input resolutions in Table~\ref{tab:compare_sdf_abc}. When increasing the input resolution, the gap of reconstruction accuracy diminishes, as reflected by CD and F1. But NMC* and NMC-lite* will always produce significantly more vertices and triangles, and take more time to process a shape.

\paragraph{Generalizability}
To show the generalizability of our method, we show the quantitative results on Thingi10K and FAUST in Table~\ref{tab:compare_sdf_thing} and ~\ref{tab:compare_sdf_faust}, respectively. Visual results can be found in Figure~\ref{fig:result_sdf} and ~\ref{fig:result_sdf_faust}. All data-driven methods are only trained on the ABC training set. These results are consistent with our analysis above.
Specifically, in terms of surface quality, we show \%IN (pred) with error greater than $5^{\circ}$ in Table~\ref{tab:compare_sdf_thing} to show that NMC* and NMC-lite* have more surface artifacts. However, on organic shapes from FAUST, MC33 outperforms deep learning methods, as shown by \%IN (pred) in Table~\ref{tab:compare_sdf_faust}. It makes sense, because Marching Cubes was originally designed to reconstruct smooth shapes, and none of the deep learning methods are trained on smooth shapes.
We also report \%SA with angles less than $10^{\circ}$ in Table~\ref{tab:compare_sdf_thing} and Table~\ref{tab:compare_sdf_faust} for Thingi10K and FAUST, to show that our method produces fewer small-angle triangles.

\subsection{Reconstruction from UDF} 
\label{sec:results_udf}
As shown in the last rows of Table ~\ref{tab:compare_sdf_abc}, ~\ref{tab:compare_sdf_thing}, and ~\ref{tab:compare_sdf_faust}, the results on UDF are similar to those on SDF, but are usually worse due to the lack of signs. Still, UNDC is able to recover the shapes reasonably well by just observing the changes of unsigned distances in nearby cells. The visual results on UDF are very similar to the results on SDF when tested on the three datasets in the previous experiment. Therefore, we show the results of reconstructing clothes with open surfaces in MGN dataset ~\cite{MGN} from grids of unsigned distances in Figure~\ref{fig:udf_clothes}. Note that in our experiments with UDF, we do not compare with prior works, since, to the best of our knowledge, UNDC is the only method that can reconstruct meshes from UDF.

\begin{table}
\caption{Quantitative results on \textbf{ABC} test set with \textbf{binary voxel} input.}
\label{tab:compare_voxel_abc}
\begin{center}
\resizebox{1.0\linewidth}{!}{
\begin{tabular}{lrrrrrrrrrrrrrrrrr}
\toprule
$64^3$ & CD$\downarrow$ & F1$\uparrow$ & NC$\uparrow$ & ECD$\downarrow$ & EF1$\uparrow$ & \#V & \#T & {\small Inference} \\
{\small Voxel input} & {\small($\times 10^5$)} & & & {\small($\times 10^2$)} & & & & {\small time} \;\; \\
\midrule
MC33        & 26.862 & 0.085 & 0.921 & 11.342 & 0.018 & 5,826 & 11,656 & {\bf 0.005s} \\
NMC*        & 9.452 & 0.422 & 0.927 & 0.698 & 0.346 & 42,045 & 84,089 & 0.156s \\
NMC-lite*   & 9.428 & 0.420 & 0.927 & 0.604 & 0.356 & 21,431 & 42,862 & 0.154s \\
NDC       & 9.387 & {\bf 0.428} & 0.930 & 0.567 & {\bf 0.360} & {\bf 5,345} & {\bf 10,726} & 0.055s \\
UNDC       & {\bf 9.139} & {\bf 0.428} & {\bf 0.931} & {\bf 0.564} & 0.359 & 5,365 & 10,772 & 0.055s \\
\bottomrule
\end{tabular}
}
\end{center}
\centering
\end{table}

\begin{figure}
  \centering
  \includegraphics[width=0.99\linewidth]{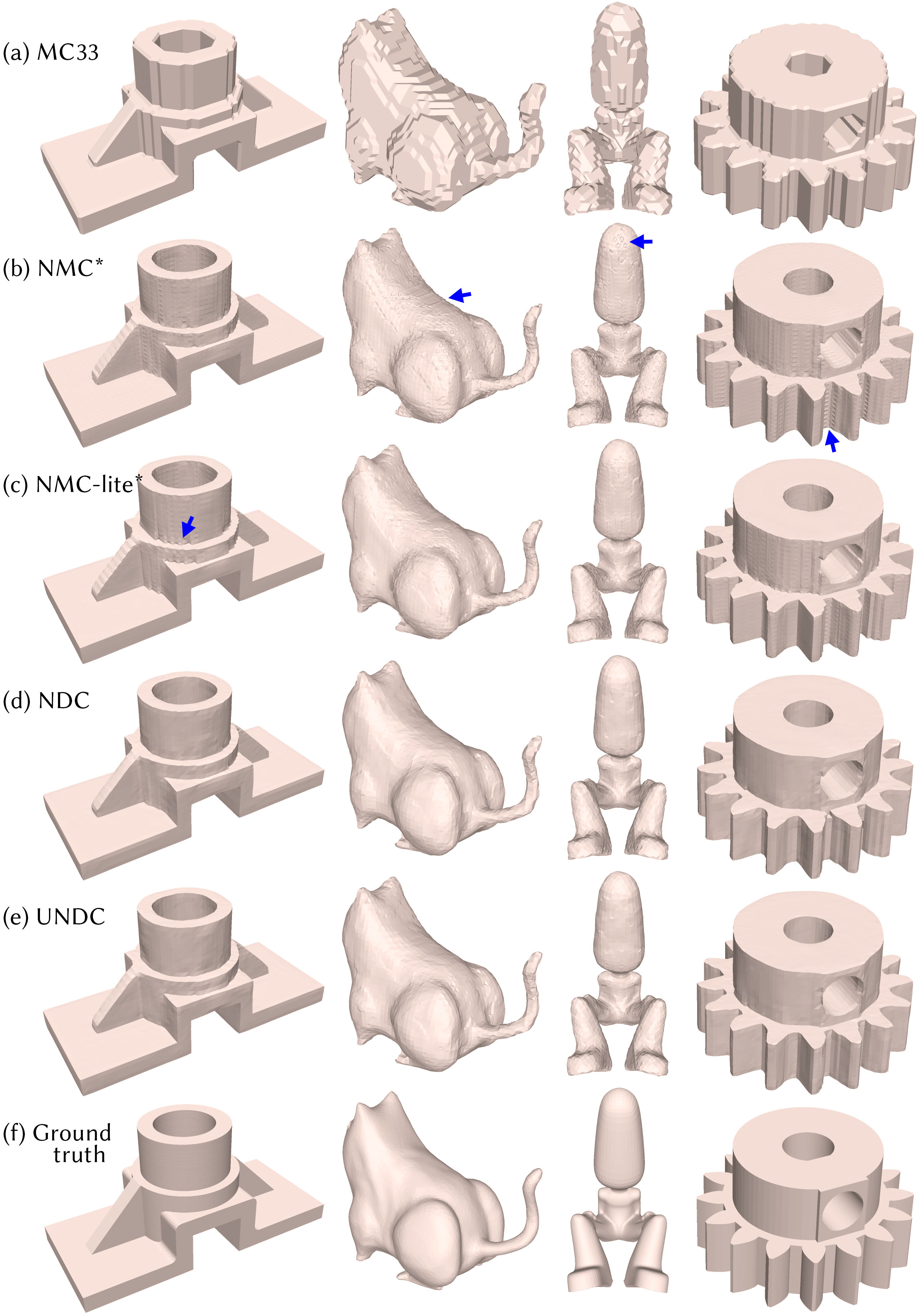}
	\caption{Mesh reconstruction results from \textbf{binary voxel} (\textbf{occupancy}) inputs at $64^3$ resolution. Zoom in to see some surface artifacts by NMC-lite* and NMC*, marked with blue arrows. The shapes in the first column are from \textbf{ABC} test set, and the last three columns from \textbf{Thingi10K}.}
	\label{fig:result_voxel}
\end{figure}

\subsection{Reconstruction from binary voxels} 
\label{sec:results_voxel}
Reconstructing meshes from binary voxels is clearly more challenging than from SDF grids. Learning from data is absolutely necessary in this scenario if one wishes to produce plausible outputs, as reflected in Table~\ref{tab:compare_voxel_abc}, where MC33 is significantly worse than all others in all metrics, except for vertex and triangle count. The results on Thingi10K and FAUST, and on $128^3$ inputs are in the \SupplementaryMaterial. They show the same pattern as Table~\ref{tab:compare_voxel_abc} and demonstrate the generalizability of our method. We show visual results in Figure~\ref{fig:result_voxel}, with more provided in the \SupplementaryMaterial.
Many observations in Section ~\ref{sec:results_sdf} still apply here: our method is significantly faster than NMC* and NMC-lite* (Inference time), produces significantly less vertices and triangles (\#V, \#T), has better normal quality (NC), and can better preserve sharp edges and corners (ECD, EF1). Moreover, since binary voxels are more challenging than SDF grids, NMC* and NMC-lite* are underfitting, with reconstruction accuracy worse than NDC and UNDC, as reflected by CD and F1.

\subsection{Reconstruction from point clouds} 
\label{sec:results_pc}
We test UNDC on the task of reconstructing meshes from point clouds.
UNDC \textit{does not} require normals as input, while most other methods do, making direct comparisons difficult to perform.
To give competing methods a slight advantage, we provide normals to methods that require them, and re-iterate that our method \textit{does not} leverage this additional information.

\begin{figure}
  \centering
  \includegraphics[width=0.99\linewidth]{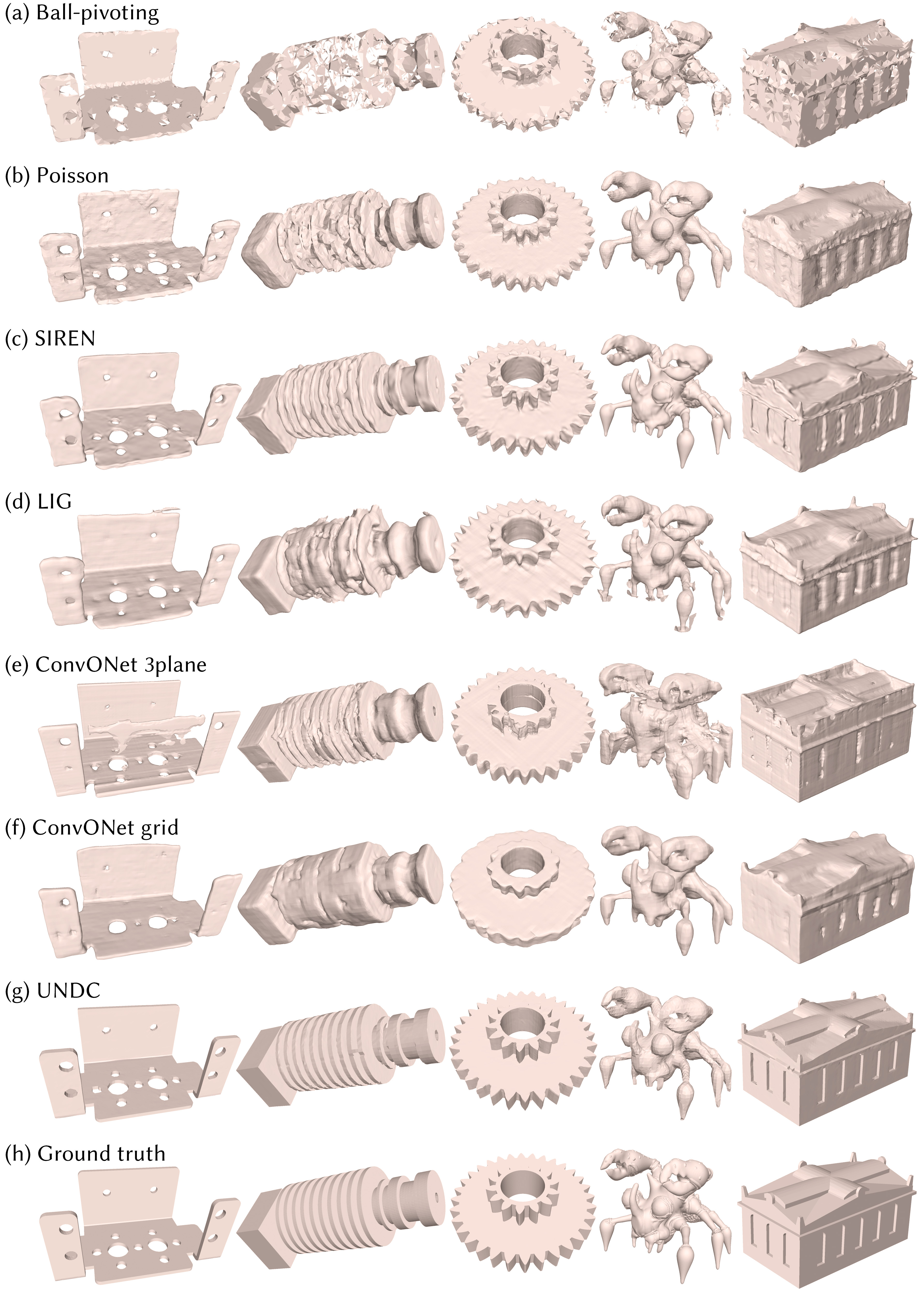}
	\caption{Results of reconstructing 3D meshes from \textbf{point cloud} inputs of 4,096 points. Please zoom in to observe the surface details. The shapes in the first two columns are from \textbf{ABC} test set, and the last three columns from \textbf{Thingi10K}. }
	\label{fig:result_pc}
\end{figure}

\paragraph{Baselines}
We compare against five methods, including classical {Ball-pivoting} \cite{bernardini1999ball}
and screened {Poisson} \cite{kazhdan2013screened}) surface reconstruction, as well as three deep learning methods like {SIREN}~\cite{sitzmann2020implicit},
Local Implicit Grids ({LIG})~\cite{jiang2020local},
and Convolutional Occupancy Networks ({ConvONet}) \cite{peng2020convolutional}.
The latter is the only method that does not require point normals, and we test its two variations proposed in the original paper:
{ConvONet-3plane} which uses the 3-plane (xy, yz, xz) setting,
and {ConvONet-grid} that uses the 3D grid setting.
Note that we compare with {SIREN} rather than {SAL}~\cite{atzmon2020sal} since {SIREN} is built upon {SAL} and has shown better performance in their paper.
We test all methods with 4,096 input points. 
The output grid size of UNDC is $64^3$.
We train all data-driven methods on the ABC training set for a fair comparison.
More detailed discussions of each method can be found in the \SupplementaryMaterial.
We illustrate these results in Figure~\ref{fig:result_pc}.
Quantitative results on the ABC test set are provided in Table~\ref{tab:compare_pointcloud_abc}, while results on Thingi10K and FAUST, which reveal a similar pattern and trend, can be found in the \SupplementaryMaterial. 

\paragraph{Analysis}
As shown in Table~\ref{tab:compare_pointcloud_abc}, UNDC outperforms all other methods in all reconstruction quality metrics.
SIREN has the closest results to ours in terms of reconstruction accuracy, but it has to be trained for, i.e., ``overfit to'', each input shape.
ConvONet, whose networks are not local, does not generalize well even within the ABC dataset.
LIG is local and therefore expected to generalize better.
However, its algorithm only considers the space around the input points and ignores the empty space.
As a result, LIG generates many artifacts in the empty space, which are called ``back-faces'' in their paper, and a post-processing step is required to remove them. The post-processing step is not perfect, as shown in the first and fourth columns in Figure~\ref{fig:result_pc}.
UNDC and Ball-pivoting are the only two methods that directly output a mesh without iso-surface extraction, therefore they have the least numbers of vertices and triangles, and are the only two methods that can generate sharp features, as shown by ECD and EF1 in Table~\ref{tab:compare_pointcloud_abc}.
As for inference time, UNDC is the fastest and significantly faster than all other methods compared, even the classical Ball-pivoting and Poisson.

\begin{table}
\caption{Quantitative results on \textbf{ABC} test set with \textbf{point cloud} input. (+n) indicates that the method additionally requires point normals as input.}
\label{tab:compare_pointcloud_abc}
\begin{center}
\resizebox{1.0\linewidth}{!}{
\begin{tabular}{lrrrrrrrr}
\toprule
point cloud & CD$\downarrow$ & F1$\uparrow$ & NC$\uparrow$ & ECD$\downarrow$ & EF1$\uparrow$ & \#V & \#T & {\small Inference} \\
(4,096) & {\small($\times 10^5$)} & & & {\small($\times 10^2$)} & & & & {\small time} \;\; \\
\midrule
{\small Ball-pivoting (+n) }   & 3.080 & 0.791 & 0.944 & 0.556 & 0.269 & {\bf 4,096} & {\bf 7,439} & 1.292s \\
Poisson  (+n)         & 4.705 & 0.727 & 0.939 & 4.138 & 0.067 & 11,241 & 22,496 & 1.476s \\
SIREN (+n)           & 1.340 & 0.814 & 0.969 & 2.636 & 0.152 & 97,219 & 194,543 & 168.595s \\
LIG (+n)             & 3.413 & 0.721 & 0.947 & 11.868 & 0.022 & 149,860 & 299,166 & 66.866s \\
{\small ConvONet 3plane} & 18.073 & 0.536 & 0.935 & 4.113 & 0.105 & 75,342 & 150,689 & 2.692s \\
{\small ConvONet grid}   & 8.844 & 0.488 & 0.939 & 9.701 & 0.036 & 74,171 & 148,337 & 2.404s \\
UNDC                & {\bf 0.893} & {\bf 0.873} & {\bf 0.974} & {\bf 0.289} & {\bf 0.757} & 5,578 & 11,261 & {\bf 0.194s}\\
\bottomrule
\end{tabular}
}
\end{center}
\centering
\end{table}

\subsection{Reconstruction from noisy real scans} 
\label{sec:results_noisy}
We test UNDC on reconstructing meshes from raw scan data in Matterport3D \cite{Matterport3D}.
The raw scan data contains depth images and camera parameters -- we convert them into noisy point clouds as the input to our network.
Since the point clouds represent large scenes, we first crop the scene into overlapping patches, and then run our network to obtain grids of edge intersection flags and vertex locations.
Finally, we put together the predicted grids to form a large grid of the scene, and then run the meshing algorithm in \Figure{undc} to obtain the output mesh.
Our network is larger than that of the previous experiment to accommodate for point cloud noise.
Specifically, we replace the three $3^3$ conv3d layers with 8 residual blocks \cite{resnet}.
During training, we also augment the input with Gaussian noise, with $\sigma=0.5$, assuming that each cell of the output grid is a unit cube, to simulate noise present in the raw scans; see \SupplementaryMaterial for more details.

\begin{figure}
  \centering
  \includegraphics[width=1.0\linewidth]{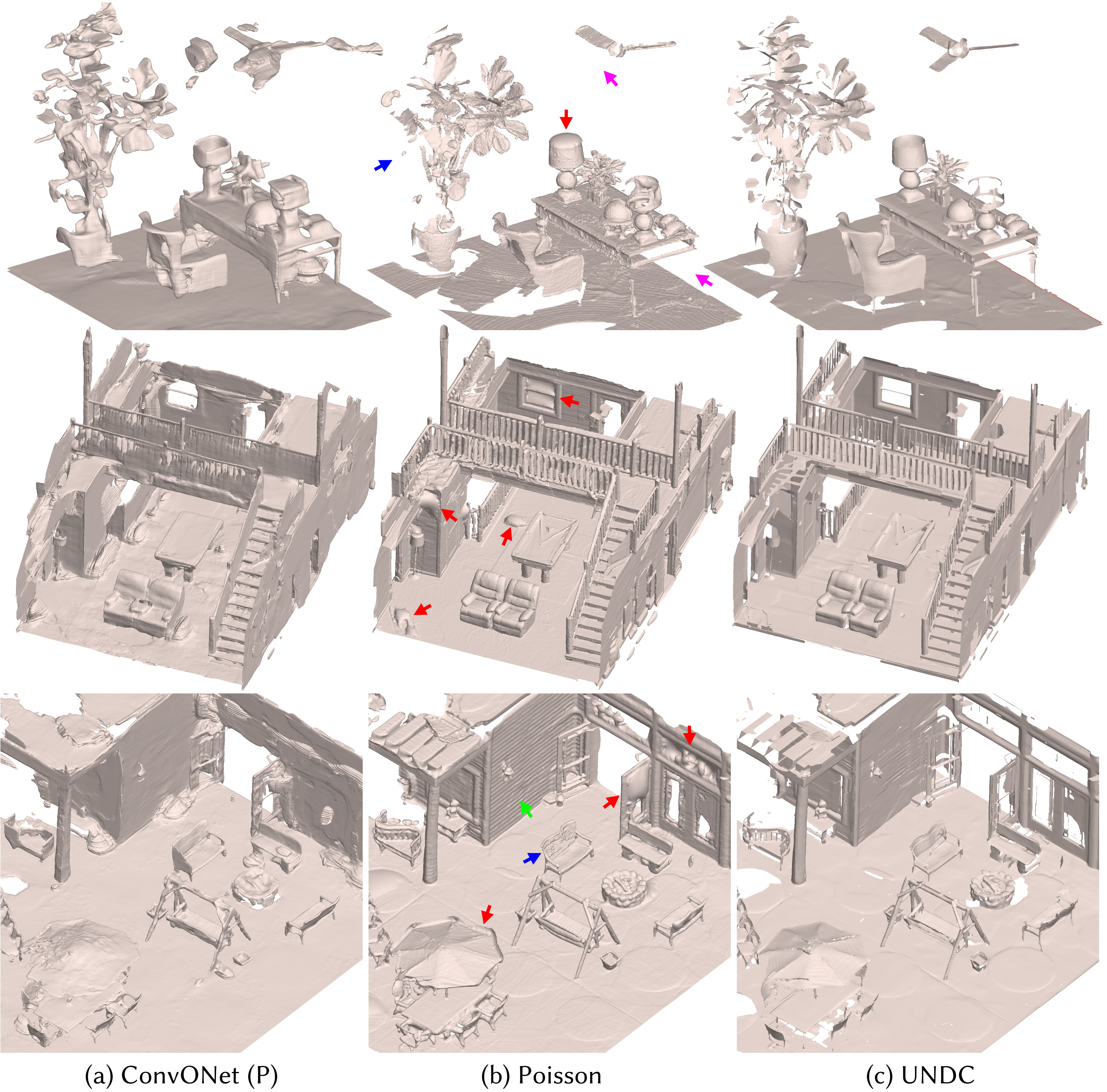}
	\caption{Qualitative comparison between {ConvONet}, {Poisson}, and \textbf{UNDC} on reconstructing rooms in \textbf{Matterport3D} from \textbf{raw scan data}, where some walls and roofs are removed for better visualization.
Colored arrows bring attention to regions where Poisson should be contrasted against UNDC.
Red arrows: ``bubble'' artifacts caused by the water-tightness prior to Poisson; purple arrows: objects or parts incorrectly trimmed; blue arrows: poor reconstruction of thin surfaces. Green arrow in the bottom row points out an instance of better preservation of surface details by Poisson (the strip patterns are {\em not\/} noise or reconstruction artifacts); the flip side of this, however, is surface noise, as seen over Poisson reconstruction in the first row.}
	\label{fig:result_scene}
\vspace{-5mm}
\end{figure}

\paragraph{Baselines}
We compare UNDC with ConvONet, since it does not require ground truth point normals and is designed to reconstruct large scenes.
We use the pre-trained weights provided by the authors for synthetic scenes with objects from ShapeNet~\cite{chang2015shapenet}, denoted as ConvONet (P).
Additionally, we compare to Poisson with estimated point normals.
We show visual comparisons in Figure~\ref{fig:result_scene}. Note that different from the experiments in most other works on deep learning scene reconstruction, which test their methods on sampled points from the ``ground truth'' meshes, we test on raw scan data, which is a more realistic setting.
We do not report quantitative results since the ``ground truth'' meshes provided with the dataset were reconstructed by Poisson, one of the methods we are comparing against.

\paragraph{Analysis}
ConvONet is seen to significantly underperform compared to Poisson and UNDC, especially falling short in terms of surface quality and detail preservation.
Therefore, we mainly compare UNDC with Poisson.
Generally, UNDC produces less surface noise, which is especially obvious in the first row of Figure~\ref{fig:result_scene}. The improvements are also observable in the other two rows, but they are less obvious due to the zoom-out to reveal the entire scenes.
Since UNDC is trained on data with noise augmentation, it learns, to some extent, to remove noise. 

Also, UNDC only reconstructs what is given in the input point cloud.
In contrast, Poisson creates an implicit field of the scene, which could potentially inpaint the missing regions.
However, such inpainting is not always desirable, and Poisson needs to trim the output mesh to remove surfaces that are generated in empty regions using a post-process (SurfaceTrimmer) that depends on careful tuning of parameters.  If the trimming density threshold is too small, it may leave ``bubble'' artifacts as indicated by red arrows in Figure~\ref{fig:result_scene}.  If it is too large, it may accidentally trim the objects, as indicated by purple arrows in Figure~\ref{fig:result_scene}.  At the default setting shown, UNDC tends to produce more holes in the output, but avoids creating bubble artifacts, see the last row of Figure~\ref{fig:result_scene}.

While water-tightness can be beneficial as a prior, it can lead to poor reconstruction of thin surfaces; this can be observed from the blue arrows in Figure~\ref{fig:result_scene}.
One thing worth special attention is the bottom-most red arrow in the last row of Figure~\ref{fig:result_scene}. The umbrella surface is thinner than a voxel.
However, Poisson forces inside-outside by creating a bubble on top of the umbrella, so that the bottom of the bubble can form the surface of the umbrella.
This creates an odd boundary after trimming.

\begin{table}
\caption{
Comparing reconstruction results of \textbf{UNDC} (output grid size at $64^3$) and Poisson on \textbf{point cloud} inputs from \textbf{ABC} test set, with varying point counts and noise levels to test the robustness of our method.
}
\label{tab:varying_density_noise}
\begin{center}
\resizebox{1.0\linewidth}{!}{
\begin{tabular}{rr|rr|rr|rr}
\toprule
Number of & Gaussian & \multicolumn{2}{|c}{CD$\downarrow$ ($\times 10^5$)} & \multicolumn{2}{|c}{F1$\uparrow$} & \multicolumn{2}{|c}{NC$\uparrow$}  \\
input points & noise levels & Poisson & UNDC & Poisson & UNDC & Poisson & UNDC \\
\midrule
1,024  & None & 34.872 & {\bf 2.510} & 0.248 & {\bf 0.806} & 0.799 & {\bf 0.944}  \\
2,048  & None & 16.406 & {\bf 1.226} & 0.387 & {\bf 0.850} & 0.847 & {\bf 0.962}  \\
4,096  & None & 8.653 & {\bf 0.987} & 0.539 & {\bf 0.867} & 0.879 & {\bf 0.970}  \\
4,096  & $\sigma=0.2$ & 9.814 & {\bf 1.179} & 0.480 & {\bf 0.840} & 0.863 & {\bf 0.962}  \\
4,096  & $\sigma=0.5$ & 14.017 & {\bf 2.061} & 0.331 & {\bf 0.717} & 0.821 & {\bf 0.935}  \\
16,384  & $\sigma=0.2$ & 5.286 & {\bf 0.936} & 0.636 & {\bf 0.866} & 0.889 & {\bf 0.971}  \\
16,384  & $\sigma=0.5$ & 8.738 & {\bf 1.236} & 0.444 & {\bf 0.813} & 0.840 & {\bf 0.955}  \\
65,536  & $\sigma=0.2$ & 2.299 & {\bf 0.905} & 0.741 & {\bf 0.872} & 0.930 & {\bf 0.973}  \\
65,536  & $\sigma=0.5$ & 4.567 & {\bf 1.079} & 0.552 & {\bf 0.836} & 0.880 & {\bf 0.962}  \\
\bottomrule
\end{tabular}
}
\end{center}
\centering
\end{table}

\begin{figure}
\centering
\includegraphics[width=0.99\linewidth]{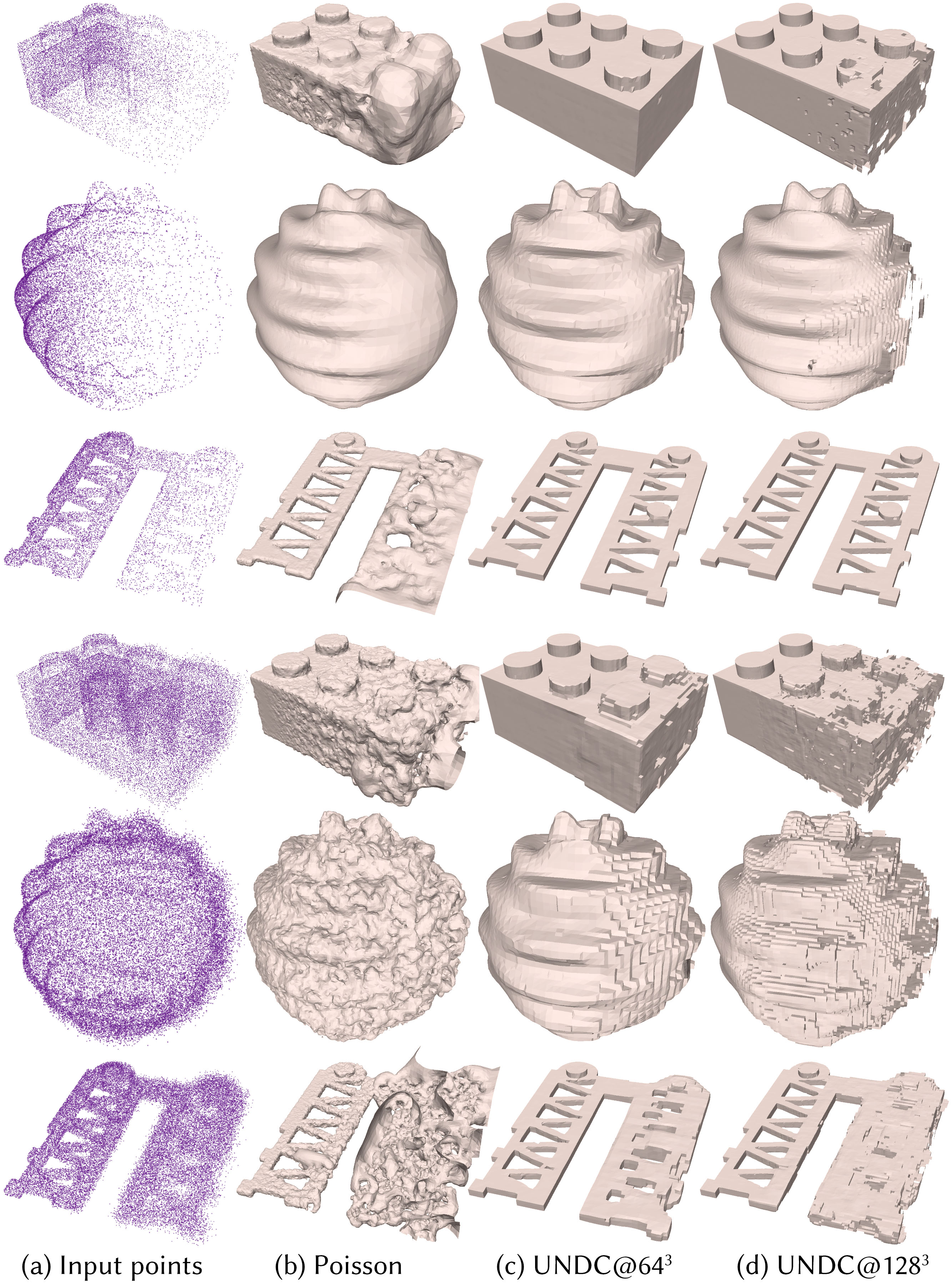}
\caption{
Qualitative comparison between {Poisson} and \textbf{UNDC} on mesh reconstruction from \textbf{point clouds} with density or noise variations \textit{across the same shape, from its left to its right, as shown}. The input point clouds in the first three rows have decreasing point density from left to right, while the inputs in the last three rows have increasing noise. UNDC@$64^3$ and UNDC@$128^3$ produce output grid sizes of $64^3$ and $128^3$, respectively.
}
\vspace{-5mm}
\label{fig:density_noise}
\end{figure}

\paragraph{Robustness to varying point density and noise}
We use synthetic data to study the robustness of UNDC on point clouds with varying density and noise. We train our network with point clouds whose point counts were randomly selected between 2,048 and 32,768, where each point cloud is augmented with Gaussian noise whose $\sigma$ is randomly sampled from $[0, 0.5]$. We then evaluate the trained network on point clouds of varying density and noise, and compare it to Poisson reconstruction with estimated point normals. 

Some quantitative results are shown in Table~\ref{tab:varying_density_noise}, where UNDC evidently outperforms Poisson. In Figure~\ref{fig:density_noise}, we show visual results where the point density or noise varies \emph{within} each input point cloud. Note that UNDC at $128^3$ output resolution tends to produce worse results than at $64^3$ output resolution. This is because relatively, point sparsity and noise level are both more significant at higher-resolution grids, due to the smaller cell sizes.

\section{Conclusions}
\label{sec:future}

We introduce neural dual contouring, a new data-driven approach to mesh reconstruction based on dual contouring.
The volumetric version of our approach, NDC, takes the same input as MC and NMC, and it can better preserve sharp features while using approximately the same number of vertices and triangles as classical MC, which is 3-7 times reduction compared to NMC. The surface version of our approach, UNDC, is sign agnostic; it is therefore able to reconstruct open surfaces and thin structures from unsigned distance fields or unoriented point clouds. Both NDC and UNDC are designed as local networks using limited receptive fields, thus can generalize well to new datasets. Extensive experiments demonstrate the superior performance of our approach on multiple datasets over state-of-the-art methods, whether learned (e.g., NMC, SIREN, LIG, ConvONet) or traditional (e.g., MC33, Poisson, Ball-Pivoting).

\begin{table}
\caption{
Statistics on non-manifold and boundary edges produced by NDC and UNDC. The methods are tested on ABC test set with $64^3$ output resolution. Non-manifold-3 denotes non-manifold edges with 3 adjacent faces, and Non-manifold-4 denotes those with 4 adjacent faces. Boundary-1 refers to boundary edges, defined as edges with only one adjacent face.}
\label{tab:open_surface_non_manifold}
\begin{center}
\resizebox{1.0\linewidth}{!}{
\begin{tabular}{lrrrrr}
\toprule
Input & \multicolumn{1}{c}{$64^3$ SDF} & \multicolumn{1}{c}{$64^3$ SDF} & \multicolumn{1}{c}{$64^3$ SDF} & \multicolumn{1}{c}{$4,096$ points} & \multicolumn{1}{c}{$4,096$ points} \\
Method & \multicolumn{1}{c}{NDC} & \multicolumn{1}{c}{UNDC} & \multicolumn{1}{c}{UNDC} & \multicolumn{1}{c}{UNDC} & \multicolumn{1}{c}{UNDC}\\
Post-processing & \multicolumn{1}{c}{No} & \multicolumn{1}{c}{No} & \multicolumn{1}{c}{Yes} & \multicolumn{1}{c}{No} & \multicolumn{1}{c}{Yes} \\
Non-manifold-3 & 0.0 (0.000\%) & 125.3 (1.116\%) & 135.4 (1.206\%) & 139.4 (1.242\%) & 168.0 (1.496\%) \\
Non-manifold-4 & 20.1 (0.183\%) & 31.2 (0.278\%) & 31.7 (0.282\%) & 28.8 (0.257\%) & 30.4 (0.271\%) \\
Boundary-1 & 0.0 (0.000\%) & 56.7 (0.505\%) & 29.6 (0.264\%) & 116.4 (1.037\%) & 41.1 (0.366\%) \\
\bottomrule
\end{tabular}
}
\end{center}
\centering
\end{table}

\paragraph{Limitations}
One limitation of our approach is that it can produce non-manifold meshes.
Specifically, since DC and its descendants produce only one vertex per grid cell, they may create meshes with vertices and edges shared by multiple surface patches in cases where MC would output multiple disconnected components within one cell; see the second column of Figure~\ref{fig:result_sdf} where the green arrows indicate non-manifold edges created by NDC and UNDC.
These cases happen fairly rarely (see statistics in Table~\ref{tab:open_surface_non_manifold}) and are easily detected and fixed by splitting vertices/edges or ``tunnelling" through them, using the techniques described in~\cite{dualmarchingcubes} or \cite{schaefer2007manifold}, for manifold dual contouring.

However, UNDC can also produce open surfaces (with edges connected to one face) or non-manifold fins (where edges are shared by three faces).
Creating open surfaces is generally good, as it allows reconstruction of thin features and partial inputs (e.g., note the thin sheets indicated by the blue arrows in the row of UNDC results in Figure~\ref{fig:result_sdf} better approximate the ground truth). However, boundaries and fins may cause problems for downstream tasks that assume manifoldness as a pre-condition. Hence, UNDC may not be the best meshing solution for all applications. 

\begin{figure}
  \centering
  \includegraphics[width=0.99\linewidth]{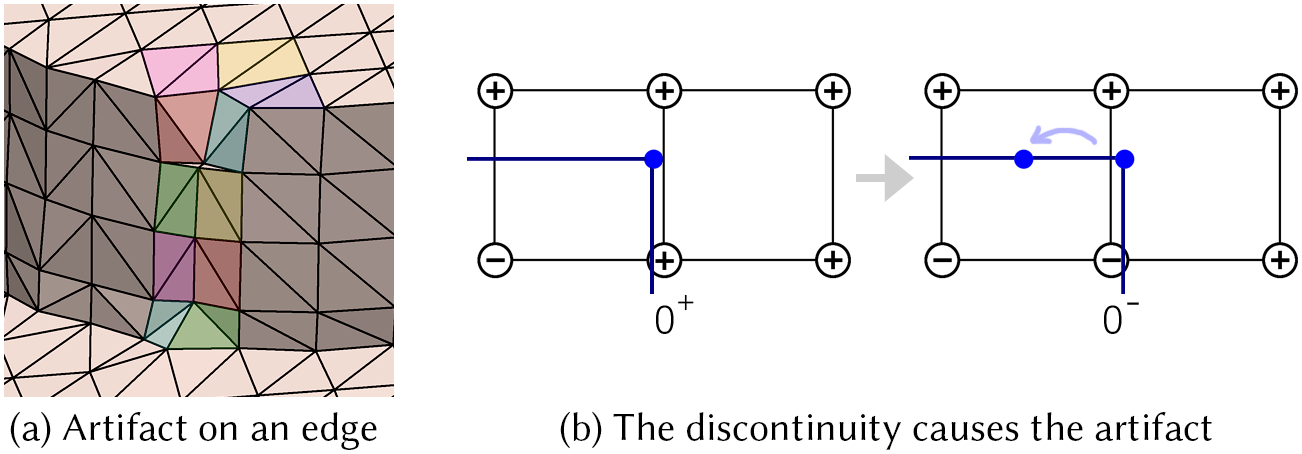}
	\caption{The edge artifacts and the cause.
	The quad faces are colored differently to show that the artifacts are not caused by \textit{random} quad splitting.
	}
	\label{fig:limitation}
\vspace{-5mm}
\end{figure}

Another limitation is that the output of NDC is not completely invariant to orientation.
Although NDC is empirically less sensitive to rotations than NMC (see \SupplementaryVideo), we still see that NDC occasionally generates  coherence artifacts on sharp edges as an object rotates. One example is shown in Figure~\ref{fig:limitation} (a). The artifact occurs when one or more vertices of the cube have SDF values very close to 0. It cannot be easily avoided since it is due to the continuity of neural networks. See the illustration in Figure~\ref{fig:limitation} (b). When the SDF value of the vertex gradually moves from positive (outside) to negative (inside), the input to the network (the SDF values) changes smoothly, but the output needs to change in a discontinuous way in order to produce the ground truth. Since most neural networks are continuous, the output of the network will be continuous. Therefore the network will generate artifacts when such transitions occur.

\paragraph{Future works}
Besides fixing the issues above, it would be interesting to incorporate the NDC framework into an end-to-end system for recovering surfaces from neural representations inferred from multiple images, possibly using Neural Radiance Fields \cite{mildenhall2020nerf}.  Or, \UNDC could potentially be used with differentiable rendering to reconstruct one-sided surfaces from a sparse set of images acquired from cameras inside a scene.   These and other NDC extensions are promising topics for future work.

\vspace{-1mm}
\begin{acks}
We thank the anonymous reviewers for their valuable comments and feedback. Thanks also go to Vincent Sitzmann, Daniel Duckworth, and Forrester Cole for helpful discussions. This work was supported in part by NSERC (no.~611370) and gift funds from Google in the form of a Google Faculty Award by the last author and a Google PhD Fellowship received by the first author.
\end{acks}
\vspace{-1mm}

\bibliographystyle{ACM-Reference-Format}
\bibliography{mainbib}

\setcounter{page}{1}
\setcounter{section}{0}

\twocolumn[
\centering
\Huge
\textbf{Neural Dual Contouring} \\
\vspace{0.5em}Supplementary Material \\
\vspace{1.0em}
]

This document provides supplemental material for ``Neural Dual Contouring.''   It contains details regarding network architectures, mask definitions, training protocols, and experimental methods.   It also has additional qualitative and quantitative results that would not fit in the main paper.

\section{Point cloud network architecture details}
\label{app:supp_cloudnet}

Our network for processing point clouds is shown in Figure~\ref{fig:network}.
The local PointNet in Figure~\ref{fig:network} is similar to the set abstraction layer in PointNet++, with the number of local clusters being the same to the number of input points.
For each point $p_i$ in the input point cloud, we find a local cluster with $K$ points, and then apply PointNet ~\cite{pointnet} using relative coordinates of those $K$ points with respect to $p_i$. In PointNet++, the local cluster is found by setting a radius $r$, so that any points whose distances to $p_i$ are smaller than $r$ will be selected into the cluster. This may bring issues such as setting appropriate $r$ values and handling situations when a cluster has too many or too few points. Therefore, we use a simpler approach to avoid the issues.
We find the $K$ nearest neighbors ($K=8$ in our experiments) of $p_i$ to form the cluster, by using a KD-tree for efficient computation.
Afterwards, we concatenate the relative coordinates of each point with its features, apply two fully-connected layers with leaky ReLU activation, and use max-pooling to aggregate the features of all $K$ points into the feature of $p_i$.
The residual block in Figure~\ref{fig:network} is a standard residual block~\cite{resnet} for fully connected layers.
The ``Pooling into grid'' module in Figure~\ref{fig:network} is essentially a local PointNet as described above. The difference is that it uses the centers of the cells in the grid as query points to find the local clusters in the input point cloud via KNN. Since obtaining the features for all cell centers in a 3D gird is very expensive ($O(N^3)$), we only compute features for the cells that are close to the input point cloud, i.e., the cells that are within 3 units (manhattan distance) to the closest point in the point cloud, assuming the size of each cell is 1 unit. All hidden layers in our network has 128 channels.
We use the same loss functions as UNDC for training the networks.

\section{Masks definitions}
\label{app:masks}

In this section, we provide definitions of $\Masks_\Signs$, $\Masks_\Vertices$, and $\Masks_\Faces$. We assume the size of each cell is 1 unit.

\paragraph{$\Masks_\Vertices$ - NDC}
For a grid cell, if its corner vertices have different signs in the ground truth SDF, we set its corresponding entry in $\Masks_\Vertices$ to 1. The other entries in $\Masks_\Vertices$ are left 0.
The definition applies for all kinds of inputs.

\paragraph{$\Masks_\Vertices$ - UNDC}
For a grid cell, if any of its edges intersects the ground truth shape, we set its corresponding entry in $\Masks_\Vertices$ to 1. The other entries in $\Masks_\Vertices$ are left 0.
The definition applies for all kinds of inputs.

\paragraph{SDF grid input - $\Masks_\Signs$ - NDC}
NDC directly use the signs of the input as $\Signs$, therefore it does not need $\Masks_\Signs$.

\paragraph{SDF and UDF grid input - $\Masks_\Faces$ - UNDC}
For an edge in a grid cell, if both of its end vertices have signed distances less than 1, we set its corresponding entry in $\Masks_\Faces$ to 1. The other entries in $\Masks_\Faces$ are left 0.

\paragraph{Binary voxel input - $\Masks_\Signs$ - NDC}
For an occupied grid cell, if it is adjacent to an unoccupied cell (in its $3^3$ local neighborhood), we set the corresponding entries for all its 8 vertices to 1. The other entries in $\Masks_\Signs$ are left 0.

\paragraph{Binary voxel input - $\Masks_\Faces$ - UNDC}
For an edge in a grid cell, if all of its four adjacent cells are occupied, we set its corresponding entry in $\Masks_\Faces$ to 1. The other entries in $\Masks_\Faces$ are left 0.

\paragraph{Point cloud input - $\Masks_\Faces$ - UNDC}
In the point cloud networks, we only compute features for the cells that are close to the input point cloud, i.e., the cells that are within 3 units (manhattan distance) to the closest point in the point cloud. Therefore, the corresponding edges stored in those cells are set to 1 in $\Masks_\Faces$. The other entries in $\Masks_\Faces$ are left 0.

\section{Training details}

Each network is trained for 400 epochs (for SDF/voxel inputs) or 250 epochs (for point cloud inputs) with a batch size of 1 (shape). We use Adam optimizer \cite{kingma2014adam} with a learning rate of $0.0001$, beta1$=0.9$ and beta2$=0.999$ for optimization. The learning rate is halved every 100 epochs. For the last experiment ``Reconstruction from noisy real scans'', we use a large point cloud network by replacing the 3 $3^3$ conv3d layers in Figure~\ref{fig:network} with 8 residual blocks \cite{resnet}. We train the network on the same ABC training set but with heavy data augmentation (random scaling and translation, in addition to the augmentations mentioned in the paper). During training, we also augment the input point clouds with Gaussian noise ($\sigma=0.5$, assuming each cell of the output grid is a unit cube) to simulate the real noise from scan data.

\section{Details of the methods in experiment ``Reconstruction from point clouds''}

\textbf{Ball-pivoting}~\cite{bernardini1999ball} and Screened Poisson surface reconstruction (\textbf{Poisson})~\cite{kazhdan2013screened}. These are classic methods for reconstructing meshes from point clouds, and they require point normals as part of the input. Ball-pivoting does not create new vertices - it only connects the existing vertices into consistently oriented triangles. Poisson constructs an implicit field according to the points and normals, then extract the surface with an octree structure. In our experiments, we use the implementation in Open3D~\cite{Open3D} for these two methods, and use a maximum depth of 8 for the octrees in Poisson.

\textbf{SIREN}~\cite{sitzmann2020implicit} is a method that overfits an neural implicit function to a given shape, therefore it takes much longer to process a shape than all other methods, since each time it needs to train a neural network from scratch. It requires point normals as part of the input. In our experiments, we use the official code released by the authors. We find that after training SIREN, there is a 10\% possibility that the output shape is covered by a shell, which cannot be easily removed since it is close to the actual shape and connected to it in many pieces. Therefore, for those shapes we have to re-train the network for several times. Nonetheless, we report the inference time in the tables assuming all shapes are successfully trained in the first go.

Local implicit grid (\textbf{LIG})~\cite{jiang2020local} is a method that first divides the input point cloud into small overlapping blocks, and then reconstruct the part in each block by optimizing a neural implicit field, and finally put the implicit fields together to reconstruct the entire shape. It requires point normals as part of the input. In our experiments, we use the official code released by the authors. The authors have released pre-trained network weights on ShapeNet~\cite{chang2015shapenet}, therefore we denote this method with pre-trained weights as \textbf{LIG (P)}. We also train this method on ABC training net for a fair comparison, denoted as \textbf{LIG}. We use $320^3$ output resolution for both models.

\begin{figure}
\centering
\includegraphics[width=\linewidth]{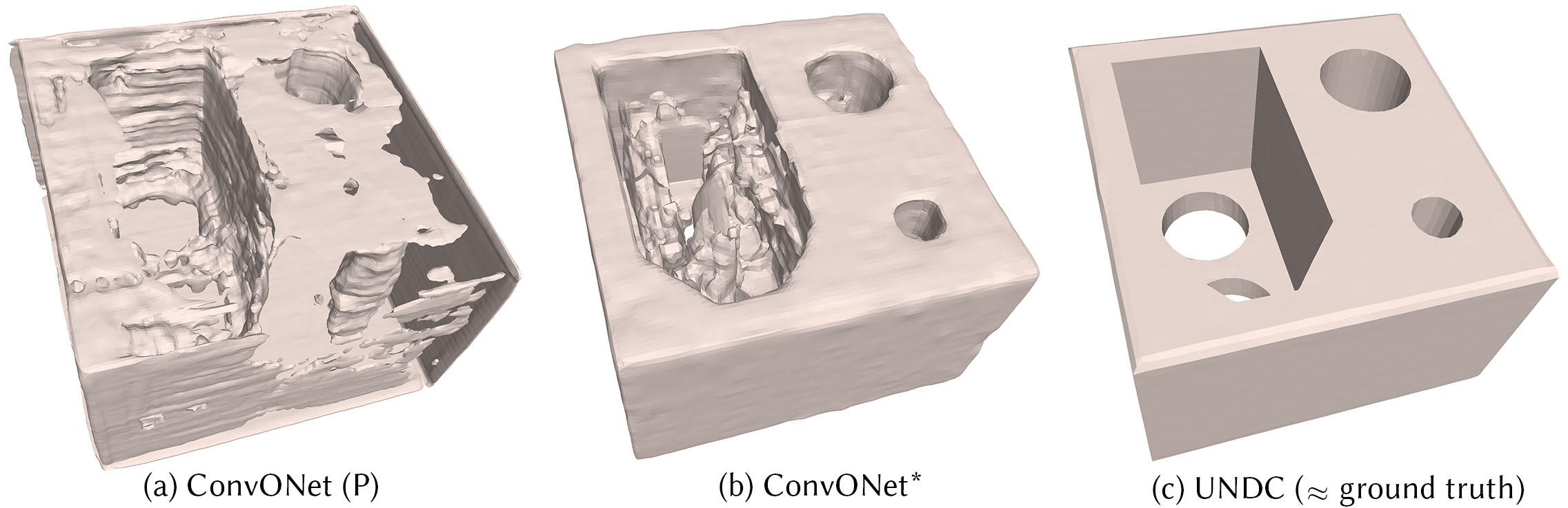}
\caption{Pre-trained ConvONet vs. ConvONet with our local backbone.}
\label{fig:local_convonet}
\end{figure}

Convolutional occupancy networks (\textbf{ConvONet})~\cite{peng2020convolutional} is a method to reconstruct an implicit field from point clouds. It does not require point normals as input, therefore is the only method that takes the exact same input as our method. In our experiments, we use the official code released by the authors. The authors have introduced many network configurations in their paper, and we choose two representative ones in our experiments. In \textbf{ConvONet 3plane}, we use the 3-plane (xy, yz, xz) setting with the resolution of each plane $128^2$. In \textbf{ConvONet grid}, we use the 3D grid setting with the grid resolution $64^3$. We train both networks on ABC training net for a fair comparison. We also use the network weights released by the authors, pre-trained on synthetic scenes with objects from ShapeNet~\cite{chang2015shapenet}, denoted as \textbf{ConvONet (P)}. We use $256^3$ output resolution for all three models. 

It is worth noting that unfortunately, all the networks in ConvONet are {\em non-local\/}, that is, their receptive fields need to cover the entire shape in order to properly decide which side is inside and which side is outside in the output implicit field. We tried to directly apply our backbone network in ConvONet, denoted as \textbf{ConvONet*}, but as expected, the training has failed. This is because our network is local, and ConvONet cannot decide inside/outside for a local patch, therefore generates many artifacts in the featureless regions, as shown in Figure~\ref{fig:local_convonet}. Note also that the pre-trained ConvONet tends to turn single-face walls into thin volumetric plates in Figure~\ref{fig:local_convonet}.

\section{Robustness to translation and rotation}

See \href{https://youtu.be/HwKMpeKgYcc}{https://youtu.be/HwKMpeKgYcc}. We test all methods on a shape while moving and rotating the shape inside the sampling grid.

\section{Additional qualitative results}

\begin{figure*}
  \centering
  \includegraphics[width=1.0\linewidth]{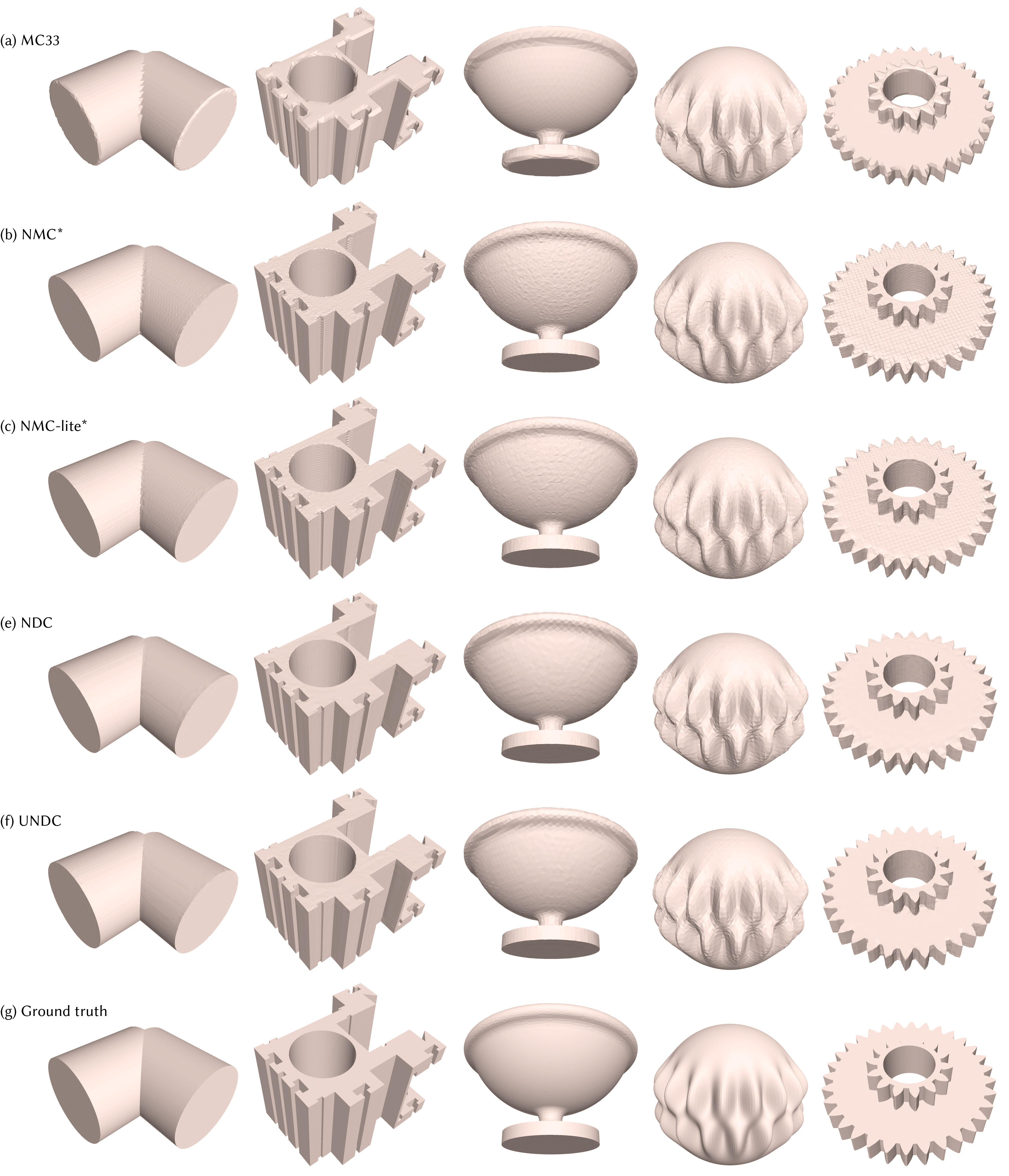}
	\caption{Results of reconstructing 3D meshes from SDF grid inputs at $64^3$ resolution. The shapes in the first three columns are from ABC test set, and the last two columns from Thingi10K. Zoom in to see the surface artifacts in NMC-lite* and NMC*.}
	\label{fig:supp_sdf}
\end{figure*}

\begin{figure*}
  \centering
  \includegraphics[width=1.0\linewidth]{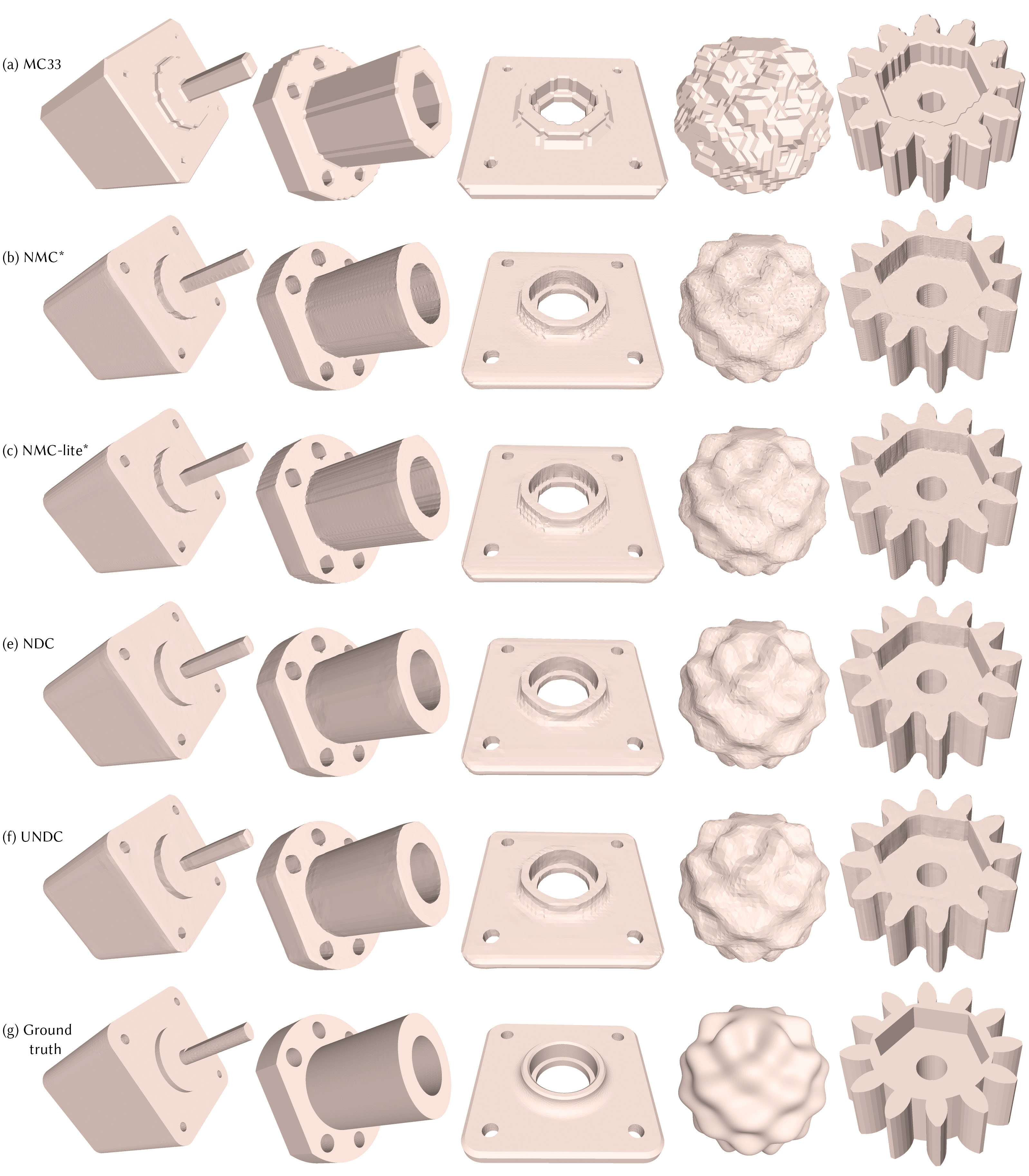}
	\caption{Results of reconstructing 3D meshes from binary voxel/occupancy inputs at $64^3$ resolution. The shapes in the first three columns are from ABC test set, and the last two columns from Thingi10K. Zoom in to see the surface artifacts in NMC-lite* and NMC*.}
	\label{fig:supp_voxel}
\end{figure*}

\begin{figure*}
  \centering
  \includegraphics[width=1.0\linewidth]{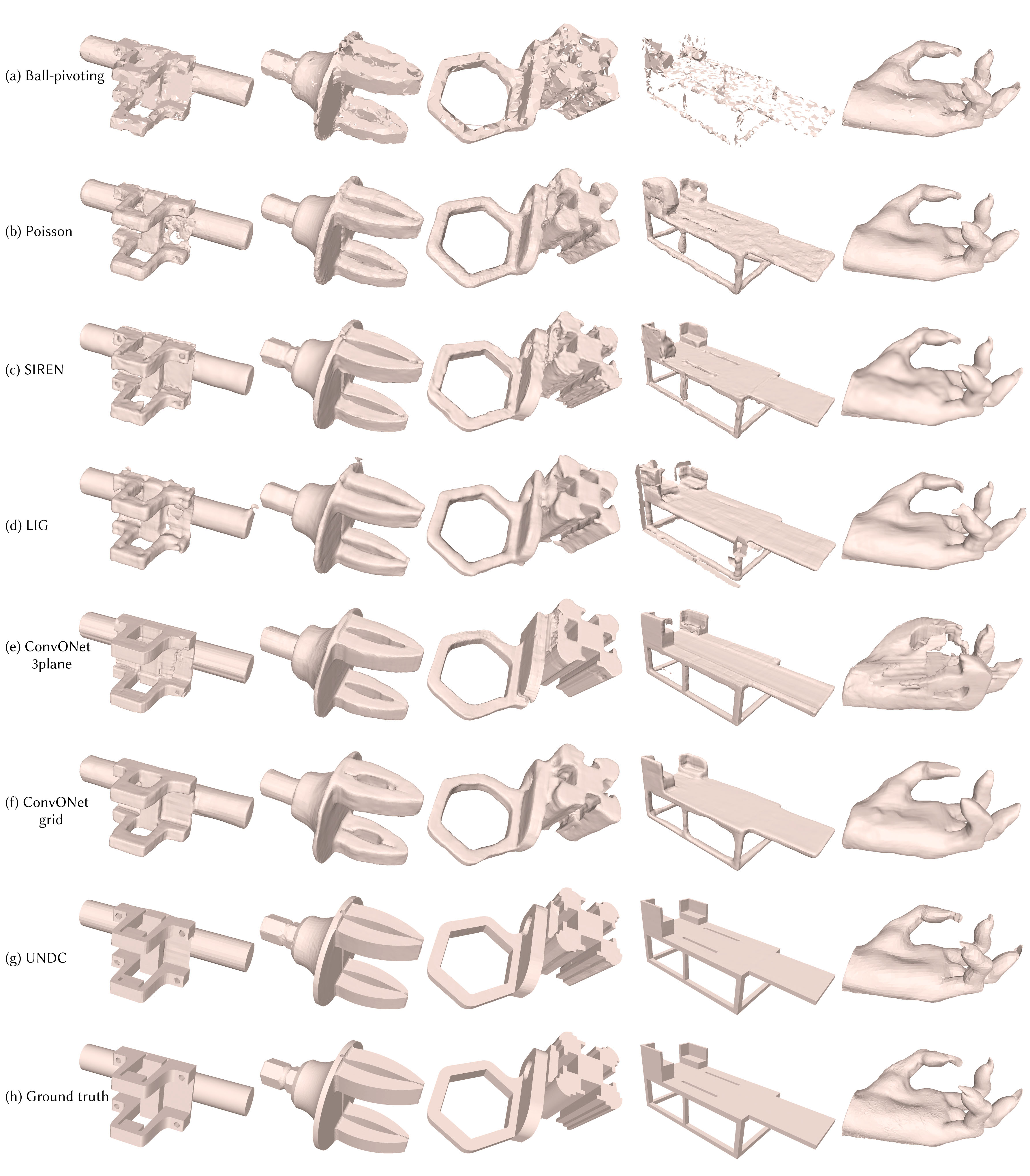}
	\caption{Results of reconstructing 3D meshes from point clouds of 4,096 points. The shapes in the first three columns are from ABC test set, and the last two columns from Thingi10K.}
	\label{fig:supp_pc}
\end{figure*}

We provide additional qualitative results via visualizations of reconstructing 3D meshes from SDF grid, binary occpuancy grid, and point cloud inputs in Figures~\ref{fig:supp_sdf}, ~\ref{fig:supp_voxel}, ~\ref{fig:supp_pc}, respectively. The input implicit fields are at $64^3$ resolution, and the input point clouds have 4,096 points each. Since the FAUST human body meshes are all very similar to the one shown in the paper, we omit FAUST results.

\section{Additional quantitative results}

We provide additional tables showing quantitative results for all combinations of input types and datasets (a subset of these tables appear in the paper due to space limitations).    
For SDF and UDF grid inputs, results for the ABC test set are in Table~\ref{tab:supp_sdf_abc}, Thingi10K in Table~\ref{tab:supp_sdf_thing}, and FAUST in Table~\ref{tab:supp_sdf_faust}.
For binary occpuancy grid inputs, the results for the ABC test set are in Table~\ref{tab:supp_voxel_abc}, Thingi10K in Table~\ref{tab:supp_voxel_thing}, and FAUST in Table~\ref{tab:supp_voxel_faust}.
For point cloud input, results for the ABC test set are in Table~\ref{tab:supp_pointcloud_abc}, Thingi10K in Table~\ref{tab:supp_pointcloud_thing}, and FAUST in Table~\ref{tab:supp_pointcloud_faust}.
In these tables, UNDC @ $64^3$ means that the output grid size of UNDC is $64^3$.

\begin{table*}
\caption{Quantitative comparison results on ABC test set with SDF and UDF grid input.}
\label{tab:supp_sdf_abc}
\begin{center}
\resizebox{1.0\linewidth}{!}{
\begin{tabular}{lrrrrrrrrrrrrrrrrr}
\toprule
$64^3$ resolution & CD$\downarrow$ & F1$\uparrow$ & NC$\uparrow$ & ECD$\downarrow$ & EF1$\uparrow$ & \#V & \#T & {\small Inference} & \multicolumn{3}{c}{\% {\footnotesize inaccurate normals (gt) }} & \multicolumn{3}{c}{\% {\footnotesize inaccurate normals (pred) }} & \multicolumn{3}{c}{\% {\small small angles}} \\
{\small SDF grid input} & {\small($\times 10^5$)} & & & {\small($\times 10^2$)} & & & & {\small time} \;\; & {\small$>80^{\circ}$} & {\small$>30^{\circ}$} & {\small$>5^{\circ}$} & {\small$>80^{\circ}$} & {\small$>30^{\circ}$} & {\small$>5^{\circ}$} & {\small$<10^{\circ}$} & {\small$<20^{\circ}$} & {\small$<30^{\circ}$} \\
\midrule
NMC         & 4.365 & 0.878 & 0.976 & 0.340 & 0.766 & 42,767 & 85,544 & 1.148s & 2.15 & 3.79 & 12.11 & 1.28 & 2.50 & 10.71 & 0.74 & 2.09 & 4.94 \\
NMC-lite    & 4.356 & 0.878 & 0.975 & 0.338 & 0.767 & 21,933 & 43,877 & 1.135s & 2.16 & 3.78 & 11.61 & 1.33 & 2.56 & 10.31 & 1.51 & 3.64 & 6.74 \\
\midrule
DC-est      & 4.673 & 0.827 & 0.958 & 3.810 & 0.167 & {\bf 5,459} & 10,969 & 0.421s & 1.68 & 9.32 & 30.17 & 0.64 & 6.62 & 27.12 & 2.29 & 6.08 & 15.24 \\
MC33        & 4.873 & 0.788 & 0.950 & 5.759 & 0.103 & 5,473 & {\bf 10,954} & {\bf 0.005s} & {\bf 1.05} & 13.48 & 31.50 & {\bf 0.40} & 9.72 & 27.03 & 1.95 & 4.09 & 6.53 \\
NMC*        & 4.400 & 0.874 & 0.972 & 0.409 & 0.715 & 42,767 & 85,544 & 0.158s & 2.01 & 4.39 & 22.03 & 1.17 & 2.98 & 20.68 & 0.57 & 1.81 & 4.63 \\
NMC-lite*   & 4.386 & {\bf 0.875} & 0.973 & 0.416 & 0.725 & 21,933 & 43,877 & 0.153s & 2.05 & 4.32 & 18.92 & 1.23 & {\bf 2.97} & 17.56 & 1.35 & 3.44 & 6.34 \\
NDC       & 4.463 & 0.867 & 0.970 & 0.338 & 0.745 & {\bf 5,459} & 10,969 & 0.027s & 2.45 & 4.66 & 16.20 & 1.48 & 3.52 & {\bf 15.03} & {\bf 0.33} & {\bf 0.76} & {\bf 4.16} \\
UNDC       & {\bf 0.930} & 0.873 & {\bf 0.974} & {\bf 0.328} & {\bf 0.746} & 5,584 & 11,295 & 0.051s & 1.65 & {\bf 3.71} & {\bf 15.75} & 1.52 & 3.69 & 15.61 & 0.44 & 0.93 & 4.41 \\
UNDC (UDF) & 0.960 & 0.868 & 0.971 & 0.379 & 0.735 & 5,692 & 11,420 & 0.053s & 1.80 & 3.93 & 16.10 & 1.70 & 4.08 & 16.21 & 0.36 & 0.89 & 4.21 \\
\toprule
\toprule
$128^3$ resolution & CD$\downarrow$ & F1$\uparrow$ & NC$\uparrow$ & ECD$\downarrow$ & EF1$\uparrow$ & \#V & \#T & {\small Inference} & \multicolumn{3}{c}{\% {\footnotesize inaccurate normals (gt) }} & \multicolumn{3}{c}{\% {\footnotesize inaccurate normals (pred) }} & \multicolumn{3}{c}{\% {\small small angles}} \\
{\small SDF grid input} & {\small($\times 10^5$)} & & & {\small($\times 10^2$)} & & & & {\small time} \;\; & {\small$>80^{\circ}$} & {\small$>30^{\circ}$} & {\small$>5^{\circ}$} & {\small$>80^{\circ}$} & {\small$>30^{\circ}$} & {\small$>5^{\circ}$} & {\small$<10^{\circ}$} & {\small$<20^{\circ}$} & {\small$<30^{\circ}$} \\
\midrule
NMC         & 4.129 & 0.882 & 0.979 & 0.204 & 0.806 & 175,926 & 351,867 & 8.991s & 2.10 & 2.98 & 8.25 & 1.26 & 1.82 & 6.97 & 0.72 & 1.84 & 4.17 \\
NMC-lite    & 4.117 & 0.882 & 0.979 & 0.231 & 0.808 & 88,419 & 176,853 & 8.984s & 2.12 & 2.97 & 7.76 & 1.28 & 1.84 & 6.53 & 1.46 & 3.35 & 5.99 \\
\midrule
DC-est      & 4.132 & 0.879 & 0.977 & 2.215 & 0.266 & 22,088 & 44,213 & 1.765s & 1.40 & 5.10 & 17.11 & 0.34 & 2.85 & 14.54 & 1.62 & 4.36 & 13.10 \\
MC33        & 4.144 & 0.870 & 0.972 & 4.247 & 0.193 & {\bf 22,048} & {\bf 44,107} & {\bf 0.030s} & {\bf 0.88} & 7.81 & 18.73 & {\bf 0.18} & 4.95 & 15.42 & 1.75 & 3.63 & 5.77 \\
NMC*        & 4.116 & 0.882 & 0.978 & 0.257 & 0.779 & 175,926 & 351,867 & 1.126s & 1.90 & 3.22 & 15.25 & 1.14 & 2.00 & 13.99 & 0.60 & 1.68 & 4.01 \\
NMC-lite*   & 4.114 & 0.882 & 0.979 & 0.283 & 0.785 & 88,419 & 176,853 & 1.112s & 1.91 & 3.15 & 12.60 & 1.18 & {\bf 1.97} & 11.35 & 1.37 & 3.26 & 5.77 \\
NDC       & 4.131 & 0.881 & 0.978 & 0.214 & 0.802 & 22,088 & 44,213 & 0.207s & 2.20 & 3.11 & 9.62 & 1.31 & 1.99 & {\bf 8.43} & {\bf 0.23} & {\bf 0.49} & {\bf 3.49} \\
UNDC       & {\bf 0.789} & {\bf 0.890} & {\bf 0.983} & {\bf 0.149} & {\bf 0.813} & 22,578 & 45,411 & 0.410s & 1.32 & {\bf 2.06} & {\bf 8.90} & 1.30 & 2.04 & 8.77 & 0.34 & 0.65 & 3.74 \\
UNDC (UDF) & 0.792 & 0.889 & {\bf 0.983} & 0.227 & 0.810 & 22,874 & 45,715 & 0.409s & 1.36 & 2.11 & 8.93 & 1.31 & 2.09 & 8.88 & {\bf 0.23} & 0.55 & 3.51 \\
\bottomrule
\end{tabular}
}
\end{center}
\centering
\end{table*}

\begin{table*}
\caption{Quantitative comparison results on Thingi10K with SDF and UDF grid input.}
\label{tab:supp_sdf_thing}
\begin{center}
\resizebox{1.0\linewidth}{!}{
\begin{tabular}{lrrrrrrrrrrrrrrrrr}
\toprule
$64^3$ resolution & CD$\downarrow$ & F1$\uparrow$ & NC$\uparrow$ & ECD$\downarrow$ & EF1$\uparrow$ & \#V & \#T & \multicolumn{3}{c}{\% {\footnotesize inaccurate normals (gt) }} & \multicolumn{3}{c}{\% {\footnotesize inaccurate normals (pred) }} & \multicolumn{3}{c}{\% {\small small angles}} \\
{\small SDF grid input} & {\small($\times 10^5$)} & & & {\small($\times 10^2$)} & & & & {\small$>80^{\circ}$} & {\small$>30^{\circ}$} & {\small$>5^{\circ}$}  & {\small$>80^{\circ}$} & {\small$>30^{\circ}$} & {\small$>5^{\circ}$}  & {\small$<10^{\circ}$} & {\small$<20^{\circ}$} & {\small$<30^{\circ}$} \\
\midrule
NMC         & 2.434 & 0.895 & 0.974 & 0.284 & 0.735 & 40,952 & 81,911 & 1.58 & 3.95 & 17.57 & 1.34 & 3.40 & 16.92 & 0.96 & 2.78 & 6.59 \\
NMC-lite    & 2.485 & 0.895 & 0.974 & 0.308 & 0.738 & 22,051 & 44,109 & 1.57 & 3.94 & 16.51 & 1.38 & 3.48 & 15.97 & 1.89 & 4.77 & 9.12 \\
\midrule
MC33        & 3.192 & 0.795 & 0.945 & 3.918 & 0.099 & 5,518 & 11,044 & {\bf 0.76} & 14.30 & 37.41 & {\bf 0.55} & 11.14 & 33.54 & 2.63 & 5.45 & 8.70 \\
NMC*        & 2.777 & {\bf 0.890} & {\bf 0.969} & 0.391 & 0.662 & 40,952 & 81,911 & 1.47 & 4.92 & 31.21 & 1.21 & 4.23 & 30.53 & 0.75 & 2.37 & 6.12 \\
NMC-lite*   & 2.760 & {\bf 0.890} & {\bf 0.969} & 0.404 & 0.674 & 22,051 & 44,109 & 1.52 & {\bf 4.82} & 27.10 & 1.28 & {\bf 4.21} & 26.42 & 1.68 & 4.47 & 8.49 \\
NDC       & 2.481 & 0.877 & 0.966 & 0.390 & {\bf 0.695} & {\bf 5,473} & {\bf 11,027} & 1.88 & 5.39 & 23.59 & 1.57 & 5.04 & {\bf 23.14} & {\bf 0.35} & {\bf 1.12} & {\bf 5.87} \\
UNDC       & {\bf 0.899} & 0.878 & 0.967 & {\bf 0.369} & 0.693 & 5,529 & 11,175 & 1.62 & 5.00 & {\bf 23.52} & 1.57 & 5.17 & 23.56 & 0.41 & 1.24 & 6.08 \\
UNDC (UDF) & 0.938 & 0.870 & 0.962 & 0.407 & 0.669 & 5,640 & 11,297 & 1.89 & 5.53 & 24.38 & 1.87 & 5.93 & 24.67 & 0.39 & 1.30 & 5.92 \\
\toprule
\toprule
$128^3$ resolution & CD$\downarrow$ & F1$\uparrow$ & NC$\uparrow$ & ECD$\downarrow$ & EF1$\uparrow$ & \#V & \#T & \multicolumn{3}{c}{\% {\footnotesize inaccurate normals (gt) }} & \multicolumn{3}{c}{\% {\footnotesize inaccurate normals (pred) }} & \multicolumn{3}{c}{\% {\small small angles}} \\
{\small SDF grid input} & {\small($\times 10^5$)} & & & {\small($\times 10^2$)} & & & & {\small$>80^{\circ}$} & {\small$>30^{\circ}$} & {\small$>5^{\circ}$}  & {\small$>80^{\circ}$} & {\small$>30^{\circ}$} & {\small$>5^{\circ}$}  & {\small$<10^{\circ}$} & {\small$<20^{\circ}$} & {\small$<30^{\circ}$} \\
\midrule
NMC         & 2.340 & 0.902 & 0.980 & 0.170 & 0.805 & 169,210 & 338,426 & 1.48 & 2.65 & 10.90 & 1.29 & 2.28 & 10.46 & 0.92 & 2.50 & 5.76 \\
NMC-lite    & 2.398 & 0.902 & 0.980 & 0.163 & 0.810 & 89,260 & 178,527 & 1.48 & 2.63 & 9.98 & 1.32 & 2.30 & 9.59 & 1.83 & 4.46 & 8.26 \\
\midrule
MC33        & 2.421 & 0.890 & 0.972 & 2.657 & 0.197 & 22,324 & 44,656 & {\bf 0.48} & 7.47 & 21.65 & {\bf 0.27} & 5.46 & 19.08 & 2.43 & 5.04 & 7.92 \\
NMC*        & 2.613 & 0.902 & 0.978 & 0.269 & 0.760 & 169,211 & 338,427 & 1.34 & 3.00 & 21.42 & 1.15 & 2.57 & 20.99 & 0.77 & 2.25 & 5.50 \\
NMC-lite*   & 2.651 & 0.902 & 0.979 & 0.254 & 0.772 & 89,260 & 178,527 & 1.37 & 2.94 & 17.49 & 1.20 & 2.54 & 17.04 & 1.74 & 4.35 & 7.91 \\
NDC       & 2.300 & 0.901 & 0.979 & 0.215 & 0.792 & {\bf 22,295} & {\bf 44,631} & 1.53 & 2.81 & 12.88 & 1.36 & 2.50 & {\bf 12.52} & {\bf 0.24} & {\bf 0.75} & {\bf 5.07} \\
UNDC       & 0.757 & {\bf 0.904} & {\bf 0.981} & {\bf 0.189} & {\bf 0.795} & 22,478 & 45,043 & 1.31 & {\bf 2.50} & {\bf 12.71} & 1.29 & {\bf 2.48} & 12.66 & 0.29 & 0.85 & 5.28 \\
UNDC (UDF) & {\bf 0.748} & 0.903 & 0.980 & 0.222 & 0.785 & 22,784 & 45,395 & 1.35 & 2.63 & 13.23 & 1.30 & 2.61 & 13.19 & 0.28 & 0.90 & 5.08 \\
\bottomrule
\end{tabular}
}
\end{center}
\centering
\end{table*}

\begin{table*}
\caption{Quantitative comparison results on FAUST with SDF and UDF grid input.}
\label{tab:supp_sdf_faust}
\begin{center}
\resizebox{1.0\linewidth}{!}{
\begin{tabular}{lrrrrrrrrrrrrrrrrr}
\toprule
$128^3$ resolution & CD$\downarrow$ & F1$\uparrow$ & NC$\uparrow$ & ECD$\downarrow$ & EF1$\uparrow$ & \#V & \#T & \multicolumn{3}{c}{\% {\footnotesize inaccurate normals (gt) }} & \multicolumn{3}{c}{\% {\footnotesize inaccurate normals (pred) }} & \multicolumn{3}{c}{\% {\small small angles}} \\
{\small SDF grid input} & {\small($\times 10^5$)} & & & {\small($\times 10^2$)} & & & & {\small$>80^{\circ}$} & {\small$>30^{\circ}$} & {\small$>5^{\circ}$}  & {\small$>80^{\circ}$} & {\small$>30^{\circ}$} & {\small$>5^{\circ}$}  & {\small$<10^{\circ}$} & {\small$<20^{\circ}$} & {\small$<30^{\circ}$} \\
\midrule
NMC         & 0.376 & 0.991 & 0.989 & 0.041 & 0.645 & 83,023 & 166,034 & 0.18 & 1.86 & 25.45 & 0.13 & 1.52 & 25.15 & 1.37 & 4.28 & 10.11 \\
NMC-lite    & 0.374 & 0.991 & 0.989 & 0.038 & 0.639 & 50,207 & 100,402 & 0.17 & 1.83 & 25.36 & 0.14 & 1.57 & 25.16 & 2.63 & 7.22 & 13.91 \\
\midrule
MC33        & 0.453 & 0.985 & 0.984 & 0.086 & 0.387 & 12,551 & {\bf 25,076} & 0.33 & 2.97 & {\bf 35.52} & {\bf 0.15} & {\bf 1.73} & {\bf 34.28} & 4.23 & 8.83 & 13.92 \\
NMC*        & 0.385 & 0.990 & 0.983 & 0.146 & 0.552 & 83,024 & 166,038 & 0.25 & 2.45 & 44.67 & 0.16 & 2.02 & 44.58 & 1.18 & 3.78 & 9.49 \\
NMC-lite*   & 0.381 & 0.991 & 0.984 & 0.119 & 0.567 & 50,207 & 100,404 & 0.22 & 2.32 & 38.51 & 0.16 & 2.00 & 38.33 & 2.63 & 7.25 & 13.60 \\
NDC       & 0.397 & 0.989 & {\bf 0.985} & 0.044 & 0.530 & {\bf 12,538} & 25,100 & 0.21 & 2.33 & 38.56 & {\bf 0.15} & 1.92 & 38.38 & {\bf 0.11} & {\bf 1.18} & {\bf 8.81} \\
UNDC       & {\bf 0.362} & {\bf 0.992} & {\bf 0.985} & {\bf 0.038} & {\bf 0.574} & 12,609 & 25,258 & {\bf 0.18} & {\bf 2.11} & 37.35 & 0.19 & 2.10 & 37.38 & 0.16 & 1.27 & 8.91 \\
UNDC (UDF) & 0.365 & 0.991 & 0.984 & 0.045 & 0.549 & 12,682 & 25,293 & 0.21 & 2.25 & 38.57 & 0.25 & 2.37 & 38.72 & 0.21 & 1.47 & 9.14 \\
\bottomrule
\end{tabular}
}
\end{center}
\centering
\end{table*}

\begin{table*}
\caption{Quantitative comparison results on ABC test set with binary occpuancy grid input.}
\label{tab:supp_voxel_abc}
\begin{center}
\resizebox{1.0\linewidth}{!}{
\begin{tabular}{lrrrrrrrrrrrrrrrrr}
\toprule
$64^3$ resolution & CD$\downarrow$ & F1$\uparrow$ & NC$\uparrow$ & ECD$\downarrow$ & EF1$\uparrow$ & \#V & \#T & {\small Inference} & \multicolumn{3}{c}{\% {\footnotesize inaccurate normals (gt) }} & \multicolumn{3}{c}{\% {\footnotesize inaccurate normals (pred) }} & \multicolumn{3}{c}{\% {\small small angles}} \\
{\small Binary voxel input} & {\small($\times 10^5$)} & & & {\small($\times 10^2$)} & & & & {\small time} \;\; & {\small$>80^{\circ}$} & {\small$>30^{\circ}$} & {\small$>5^{\circ}$} & {\small$>80^{\circ}$} & {\small$>30^{\circ}$} & {\small$>5^{\circ}$} & {\small$<10^{\circ}$} & {\small$<20^{\circ}$} & {\small$<30^{\circ}$} \\
\midrule
NMC         & 9.327 & 0.440 & 0.930 & 0.546 & 0.373 & 42,044 & 84,088 & 0.715s & 6.24 & 9.84 & 28.57 & 4.56 & 7.88 & 26.89 & 0.08 & 0.55 & 2.23 \\
NMC-lite    & 9.285 & 0.440 & 0.929 & 0.562 & 0.373 & 21,457 & 42,916 & 0.729s & 6.32 & 9.95 & 27.99 & 4.68 & 8.09 & 26.38 & 0.14 & 1.43 & 4.38 \\
\midrule
MC33        & 26.862 & 0.085 & 0.921 & 11.342 & 0.018 & 5,826 & 11,656 & {\bf 0.005s} & {\bf 4.51} & 17.42 & 40.09 & {\bf 1.47} & 16.07 & 40.27 & {\bf 0.00} & {\bf 0.00} & {\bf 1.21} \\
NMC*        & 9.452 & 0.422 & 0.927 & 0.698 & 0.346 & 42,045 & 84,089 & 0.156s & 6.25 & 10.53 & 33.14 & 4.47 & 8.45 & 31.44 & 0.02 & 0.21 & 1.52 \\
NMC-lite*   & 9.428 & 0.420 & 0.927 & 0.604 & 0.356 & 21,431 & 42,862 & 0.154s & 6.35 & 10.46 & 31.08 & 4.58 & 8.43 & 29.38 & 0.11 & 1.09 & 4.13 \\
NDC       & 9.387 & {\bf 0.428} & 0.930 & 0.567 & {\bf 0.360} & {\bf 5,345} & {\bf 10,726} & 0.055s & 6.14 & 10.11 & {\bf 29.21} & 4.39 & 8.00 & {\bf 27.25} & 0.21 & 0.38 & 2.52 \\
UNDC       & {\bf 9.139} & {\bf 0.428} & {\bf 0.931} & {\bf 0.564} & 0.359 & 5,365 & 10,772 & 0.055s & 6.02 & {\bf 9.94} & 29.51 & 4.36 & {\bf 7.98} & 27.65 & 0.21 & 0.39 & 2.53 \\
\toprule
\toprule
$128^3$ resolution & CD$\downarrow$ & F1$\uparrow$ & NC$\uparrow$ & ECD$\downarrow$ & EF1$\uparrow$ & \#V & \#T & {\small Inference} & \multicolumn{3}{c}{\% {\footnotesize inaccurate normals (gt) }} & \multicolumn{3}{c}{\% {\footnotesize inaccurate normals (pred) }} & \multicolumn{3}{c}{\% {\small small angles}} \\
{\small Binary voxel input} & {\small($\times 10^5$)} & & & {\small($\times 10^2$)} & & & & {\small time} \;\; & {\small$>80^{\circ}$} & {\small$>30^{\circ}$} & {\small$>5^{\circ}$} & {\small$>80^{\circ}$} & {\small$>30^{\circ}$} & {\small$>5^{\circ}$} & {\small$<10^{\circ}$} & {\small$<20^{\circ}$} & {\small$<30^{\circ}$} \\
\midrule
NMC         & 5.447 & 0.663 & 0.959 & 0.410 & 0.692 & 174,257 & 348,519 & 5.405s & 3.91 & 5.74 & 22.16 & 2.50 & 4.12 & 20.76 & 0.10 & 0.67 & 2.19 \\
NMC-lite    & 5.444 & 0.663 & 0.958 & 0.417 & 0.693 & 87,419 & 174,844 & 5.449s & 3.96 & 5.81 & 21.69 & 2.56 & 4.25 & 20.36 & 0.20 & 1.69 & 4.60 \\
\midrule
MC33        & 9.800 & 0.212 & 0.944 & 11.690 & 0.023 & 22,775 & 45,557 & {\bf 0.030s} & {\bf 2.71} & 12.52 & 36.74 & {\bf 1.01} & 11.78 & 37.10 & {\bf 0.00} & {\bf 0.00} & {\bf 0.96} \\
NMC*        & 5.465 & 0.659 & 0.956 & 0.652 & 0.664 & 174,255 & 348,515 & 1.129s & 3.90 & 6.30 & 27.20 & 2.44 & 4.64 & 25.88 & 0.02 & 0.27 & 1.51 \\
NMC-lite*   & 5.460 & 0.658 & 0.957 & 0.398 & 0.685 & 87,369 & 174,743 & 1.125s & 3.93 & 6.20 & 25.23 & 2.48 & 4.56 & 23.87 & 0.15 & 1.31 & 4.40 \\
NDC       & {\bf 5.451} & {\bf 0.661} & 0.960 & 0.316 & {\bf 0.686} & {\bf 21,848} & {\bf 43,715} & 0.403s & 3.85 & 5.69 & {\bf 21.01} & 2.40 & 3.97 & {\bf 19.42} & 0.13 & 0.25 & 2.55 \\
UNDC       & 5.458 & {\bf 0.661} & {\bf 0.961} & {\bf 0.315} & 0.682 & 21,877 & 43,757 & 0.404s & 3.80 & {\bf 5.63} & 21.31 & 2.37 & {\bf 3.93} & 19.73 & 0.14 & 0.26 & 2.54 \\
\bottomrule
\end{tabular}
}
\end{center}
\centering
\end{table*}

\begin{table*}
\caption{Quantitative comparison results on Thingi10K with binary occpuancy grid input.}
\label{tab:supp_voxel_thing}
\begin{center}
\resizebox{1.0\linewidth}{!}{
\begin{tabular}{lrrrrrrrrrrrrrrrrr}
\toprule
$64^3$ resolution & CD$\downarrow$ & F1$\uparrow$ & NC$\uparrow$ & ECD$\downarrow$ & EF1$\uparrow$ & \#V & \#T & \multicolumn{3}{c}{\% {\footnotesize inaccurate normals (gt) }} & \multicolumn{3}{c}{\% {\footnotesize inaccurate normals (pred) }} & \multicolumn{3}{c}{\% {\small small angles}} \\
{\small Binary voxel input} & {\small($\times 10^5$)} & & & {\small($\times 10^2$)} & & & & {\small$>80^{\circ}$} & {\small$>30^{\circ}$} & {\small$>5^{\circ}$}  & {\small$>80^{\circ}$} & {\small$>30^{\circ}$} & {\small$>5^{\circ}$}  & {\small$<10^{\circ}$} & {\small$<20^{\circ}$} & {\small$<30^{\circ}$} \\
\midrule
NMC         & 6.081 & 0.491 & 0.922 & 0.573 & 0.342 & 40,431 & 80,862 & 5.42 & 11.57 & 41.64 & 4.63 & 10.39 & 40.70 & 0.15 & 0.87 & 3.10 \\
NMC-lite    & 6.056 & 0.490 & 0.920 & 0.604 & 0.341 & 21,635 & 43,272 & 5.54 & 11.71 & 40.44 & 4.83 & 10.71 & 39.65 & 0.22 & 1.91 & 5.93 \\
\midrule
MC33        & 25.523 & 0.069 & 0.907 & 7.542 & 0.017 & 5,940 & 11,882 & {\bf 3.86} & 21.40 & 52.50 & {\bf 1.64} & 20.06 & 52.99 & {\bf 0.00} & {\bf 0.00} & 2.30 \\
NMC*        & 6.256 & 0.471 & 0.916 & 0.772 & 0.306 & 40,382 & 80,764 & 5.49 & 13.03 & 46.36 & 4.53 & 11.69 & 45.36 & 0.03 & 0.34 & {\bf 2.13} \\
NMC-lite*   & 6.226 & 0.471 & 0.917 & {\bf 0.625} & 0.321 & 21,577 & 43,154 & 5.58 & 12.79 & 44.13 & 4.64 & 11.53 & 43.18 & 0.17 & 1.46 & 5.51 \\
NDC       & 6.185 & 0.477 & 0.921 & 0.681 & {\bf 0.322} & {\bf 5,373} & {\bf 10,808} & 5.29 & 12.03 & {\bf 42.19} & 4.44 & 10.80 & {\bf 41.00} & 0.19 & 0.49 & 3.78 \\
UNDC       & {\bf 6.070} & {\bf 0.478} & {\bf 0.923} & 0.651 & 0.321 & 5,401 & 10,855 & 5.08 & {\bf 11.78} & 42.52 & 4.36 & {\bf 10.76} & 41.50 & 0.20 & 0.51 & 3.81 \\
\toprule
\toprule
$128^3$ resolution & CD$\downarrow$ & F1$\uparrow$ & NC$\uparrow$ & ECD$\downarrow$ & EF1$\uparrow$ & \#V & \#T & \multicolumn{3}{c}{\% {\footnotesize inaccurate normals (gt) }} & \multicolumn{3}{c}{\% {\footnotesize inaccurate normals (pred) }} & \multicolumn{3}{c}{\% {\small small angles}} \\
{\small Binary voxel input} & {\small($\times 10^5$)} & & & {\small($\times 10^2$)} & & & & {\small$>80^{\circ}$} & {\small$>30^{\circ}$} & {\small$>5^{\circ}$}  & {\small$>80^{\circ}$} & {\small$>30^{\circ}$} & {\small$>5^{\circ}$}  & {\small$<10^{\circ}$} & {\small$<20^{\circ}$} & {\small$<30^{\circ}$} \\
\midrule
NMC         & 3.162 & 0.726 & 0.957 & 0.404 & 0.645 & 168,218 & 336,440 & 2.95 & 6.08 & 32.64 & 2.53 & 5.37 & 32.07 & 0.19 & 1.06 & 3.23 \\
NMC-lite    & 3.163 & 0.726 & 0.956 & 0.414 & 0.650 & 88,499 & 177,003 & 2.99 & 6.17 & 31.55 & 2.59 & 5.55 & 31.07 & 0.32 & 2.33 & 6.33 \\
\midrule
MC33        & 8.473 & 0.169 & 0.934 & 7.328 & 0.026 & 23,198 & 46,400 & {\bf 1.82} & 15.34 & 48.95 & {\bf 1.09} & 15.21 & 49.98 & {\bf 0.00} & {\bf 0.00} & {\bf 2.04} \\
NMC*        & {\bf 3.184} & 0.721 & 0.951 & 0.654 & 0.604 & 168,118 & 336,240 & 2.95 & 7.25 & 38.76 & 2.46 & 6.45 & 38.22 & 0.04 & 0.44 & 2.28 \\
NMC-lite*   & 3.197 & 0.721 & 0.953 & 0.435 & 0.638 & 88,411 & 176,826 & 2.97 & 6.98 & 36.21 & 2.49 & 6.23 & 35.65 & 0.24 & 1.81 & 6.04 \\
NDC       & 3.192 & {\bf 0.724} & 0.959 & 0.427 & {\bf 0.643} & {\bf 22,109} & {\bf 44,246} & 2.85 & 5.94 & {\bf 30.72} & 2.39 & 5.12 & {\bf 29.90} & 0.14 & 0.37 & 3.88 \\
UNDC       & 3.205 & {\bf 0.724} & {\bf 0.960} & {\bf 0.369} & 0.639 & 22,157 & 44,318 & 2.74 & {\bf 5.83} & 31.13 & 2.34 & {\bf 5.09} & 30.35 & 0.14 & 0.37 & 3.84 \\
\bottomrule
\end{tabular}
}
\end{center}
\centering
\end{table*}

\begin{table*}
\caption{Quantitative comparison results on FAUST with binary occpuancy grid input.}
\label{tab:supp_voxel_faust}
\begin{center}
\resizebox{1.0\linewidth}{!}{
\begin{tabular}{lrrrrrrrrrrrrrrrrr}
\toprule
$128^3$ resolution & CD$\downarrow$ & F1$\uparrow$ & NC$\uparrow$ & ECD$\downarrow$ & EF1$\uparrow$ & \#V & \#T & \multicolumn{3}{c}{\% {\footnotesize inaccurate normals (gt) }} & \multicolumn{3}{c}{\% {\footnotesize inaccurate normals (pred) }} & \multicolumn{3}{c}{\% {\small small angles}} \\
{\small Binary voxel input} & {\small($\times 10^5$)} & & & {\small($\times 10^2$)} & & & & {\small$>80^{\circ}$} & {\small$>30^{\circ}$} & {\small$>5^{\circ}$}  & {\small$>80^{\circ}$} & {\small$>30^{\circ}$} & {\small$>5^{\circ}$}  & {\small$<10^{\circ}$} & {\small$<20^{\circ}$} & {\small$<30^{\circ}$} \\
\midrule
NMC         & 0.760 & 0.970 & 0.965 & 0.328 & 0.347 & 82,406 & 164,806 & 0.69 & 5.47 & 70.31 & 0.29 & 4.29 & 70.06 & 0.34 & 2.07 & 6.55 \\
NMC-lite    & 0.754 & 0.970 & 0.964 & 0.296 & 0.334 & 49,699 & 99,393 & 0.70 & 5.79 & 69.04 & 0.32 & 4.79 & 68.87 & 0.48 & 4.09 & 11.54 \\
\midrule
MC33        & 6.928 & 0.064 & 0.905 & 0.455 & 0.085 & 14,175 & 28,348 & {\bf 0.52} & 25.51 & 95.40 & {\bf 0.10} & 22.46 & 95.46 & {\bf 0.00} & {\bf 0.00} & 5.23 \\
NMC*        & 0.816 & 0.967 & 0.944 & 0.760 & 0.160 & 82,337 & 164,668 & 1.02 & 9.46 & 79.88 & 0.55 & 8.49 & 80.00 & 0.07 & 0.82 & {\bf 4.44} \\
NMC-lite*   & 0.801 & 0.967 & 0.953 & 0.345 & 0.311 & 49,643 & 99,279 & 0.75 & 8.27 & 75.80 & 0.29 & 7.26 & 75.75 & 0.39 & 3.21 & 10.85 \\
NDC       & 0.766 & {\bf 0.969} & {\bf 0.966} & {\bf 0.169} & 0.330 & {\bf 12,411} & {\bf 24,833} & 0.66 & 5.45 & {\bf 67.71} & 0.31 & {\bf 4.23} & {\bf 67.38} & 0.03 & 0.63 & 8.07 \\
UNDC       & {\bf 0.760} & {\bf 0.969} & {\bf 0.966} & 0.177 & {\bf 0.353} & 12,467 & 24,930 & 0.59 & {\bf 5.30} & 68.22 & 0.33 & 4.39 & 68.01 & 0.05 & 0.62 & 8.16 \\
\bottomrule
\end{tabular}
}
\end{center}
\centering
\end{table*}

\begin{table*}
\caption{Quantitative results on ABC test set with point cloud input. (+n) indicates that the method additionally requires point normals as input.}
\label{tab:supp_pointcloud_abc}
\begin{center}
\begin{tabular}{lrrrrrrrr}
\toprule
point cloud & CD$\downarrow$ & F1$\uparrow$ & NC$\uparrow$ & ECD$\downarrow$ & EF1$\uparrow$ & \#V & \#T & {\small Inference} \\
(4,096) & {\small($\times 10^5$)} & & & {\small($\times 10^2$)} & & & & {\small time} \;\; \\
\midrule
{\small Ball-pivoting (+n) }   & 3.080 & 0.791 & 0.944 & 0.556 & 0.269 & {\bf 4,096} & {\bf 7,439} & 1.292s \\
Poisson  (+n)         & 4.705 & 0.727 & 0.939 & 4.138 & 0.067 & 11,241 & 22,496 & 1.476s \\
SIREN (+n)           & 1.340 & 0.814 & 0.969 & 2.636 & 0.152 & 97,219 & 194,543 & 168.595s \\
LIG (P) (+n)         & 4.747 & 0.709 & 0.939 & 10.786 & 0.023 & 148,927 & 297,766 & 61.176s \\
LIG (+n)             & 3.413 & 0.721 & 0.947 & 11.868 & 0.022 & 149,860 & 299,166 & 66.866s \\
{\small ConvONet (P)}    & 38.926 & 0.207 & 0.844 & 1.522 & 0.057 &  127,247 & 254,627 & 4.598s \\
{\small ConvONet 3plane} & 18.073 & 0.536 & 0.935 & 4.113 & 0.105 & 75,342 & 150,689 & 2.692s \\
{\small ConvONet grid}   & 8.844 & 0.488 & 0.939 & 9.701 & 0.036 & 74,171 & 148,337 & 2.404s \\
UNDC @ $64^3$        & {\bf 0.893} & {\bf 0.873} & {\bf 0.974} & {\bf 0.289} & {\bf 0.757} & 5,578 & 11,261 & {\bf 0.194s}\\
\bottomrule
\end{tabular}
\end{center}
\centering
\end{table*}

\begin{table*}
\caption{Quantitative results on Thingi10K with point cloud input. (+n) indicates that the method additionally requires point normals as input.}
\label{tab:supp_pointcloud_thing}
\begin{center}
\begin{tabular}{lrrrrrrr}
\toprule
point cloud & CD$\downarrow$ & F1$\uparrow$ & NC$\uparrow$ & ECD$\downarrow$ & EF1$\uparrow$ & \#V & \#T \\
(4,096) & {\small($\times 10^5$)} & & & {\small($\times 10^2$)} & & & \\
\midrule
{\small Ball-pivoting (+n) }   & 2.329 & 0.787 & 0.936 & 0.602 & 0.236 & {\bf 4,096} & {\bf 7,455} \\
Poisson  (+n)         & 12.799 & 0.744 & 0.938 & 3.439 & 0.052 & 11,498 & 23,010 \\
SIREN (+n)           & 1.419 & 0.834 & 0.962 & 2.059 & 0.144 & 94,797 & 189,637 \\
LIG (P) (+n)         & 4.453 & 0.691 & 0.929 & 8.471 & 0.019 & 146,554 & 292,847 \\
LIG (+n)             & 5.991 & 0.748 & 0.943 & 8.266 & 0.021 & 145,269 & 290,354\\
{\small ConvONet (P)}    & 39.822 & 0.209 & 0.826 & 1.306 & 0.051 & 120,475 & 241,068 \\
{\small ConvONet 3plane} & 18.272 & 0.484 & 0.903 & 3.222 & 0.090 & 74,514 & 149,033 \\
{\small ConvONet grid}   & 6.032 & 0.476 & 0.928 & 8.249 & 0.024 & 73,745 & 147,484 \\
UNDC @ $64^3$      & {\bf 0.927} & {\bf 0.873} & {\bf 0.965} & {\bf 0.400} & {\bf 0.686} & 5,543 & 11,159 \\
\bottomrule
\end{tabular}
\end{center}
\centering
\end{table*}

\begin{table*}
\caption{Quantitative results on FAUST with point cloud input. (+n) indicates that the method additionally requires point normals as input.}
\label{tab:supp_pointcloud_faust}
\begin{center}
\begin{tabular}{lrrrrrrr}
\toprule
point cloud & CD$\downarrow$ & F1$\uparrow$ & NC$\uparrow$ & ECD$\downarrow$ & EF1$\uparrow$ & \#V & \#T \\
(4,096) & {\small($\times 10^5$)} & & & {\small($\times 10^2$)} & & & \\
\midrule
{\small Ball-pivoting (+n) }   & 0.906 & 0.932 & 0.965 & 0.372 & 0.131 & 4,096 & 7,652 \\
Poisson  (+n)         & 0.724 & 0.966 & 0.975 & 0.467 & 0.246 & 11,330 & 22,646 \\
SIREN (+n)           & 0.697 & 0.951 & {\bf 0.986} & 0.211 & {\bf 0.504} & 51,215 & 102,443 \\
LIG (P) (+n)         & 1.449 & 0.876 & 0.964 & 1.307 & 0.107 & 79,337 & 158,605 \\
LIG (+n)             & 2.533 & 0.871 & 0.962 & 1.772 & 0.077 & 80,845 & 161,226 \\
{\small ConvONet (P)}    & 17.334 & 0.312 & 0.849 & 1.244 & 0.027 & 47,716 & 95,424 \\
{\small ConvONet 3plane} & 23.809 & 0.389 & 0.868 & 1.046 & 0.053 & 46,211 & 92,427 \\
{\small ConvONet grid}   & 3.506 & 0.574 & 0.945 & 4.618 & 0.029 & 41,710 & 83,418 \\
UNDC @ $64^3$     & 0.532 & 0.970 & 0.965 & 0.345 & 0.206 & {\bf 3,146} & {\bf 6,308} \\
UNDC @ $128^3$    & {\bf 0.413} & {\bf 0.985} & 0.978 & {\bf 0.095} & 0.437 & 12,681 & 25,202\\
\toprule
\toprule
point cloud & CD$\downarrow$ & F1$\uparrow$ & NC$\uparrow$ & ECD$\downarrow$ & EF1$\uparrow$ & \#V & \#T \\
(16,384) & {\small($\times 10^5$)} & & & {\small($\times 10^2$)} & & & \\
\midrule
{\small Ball-pivoting (+n) }   & 0.545 & 0.977 & 0.977 & 0.144 & 0.316 & 16,384 & 31,767 \\
Poisson  (+n)        & 0.397 & 0.987 & 0.987 & 0.118 & 0.528 & 45,325 & 90,630 \\
SIREN (+n)           & 0.707 & 0.953 & 0.988 & 0.263 & 0.562 & 51,132 & 102,270 \\
LIG (P) (+n)         & 1.140 & 0.902 & 0.969 & 1.170 & 0.160 & 78,821 & 157,622\\
LIG (+n)             & 2.215 & 0.895 & 0.966 & 1.792 & 0.120 & 80,399 & 160,445 \\
{\small ConvONet (P)}    & 20.218 & 0.250 & 0.865 & 1.345 & 0.024 & 51,449 & 102,873 \\
{\small ConvONet 3plane} & 24.682 & 0.390 & 0.869 & 0.985 & 0.048 & 47,922 & 95,851 \\
{\small ConvONet grid}   & 3.563 & 0.568 & 0.947 & 5.474 & 0.029 & 42,218 & 84,433 \\
UNDC @ $128^3$     & 0.368 & 0.991 & 0.983 & 0.050 & 0.566 & {\bf 12,665} & {\bf 25,387} \\
UNDC @ $256^3$     & {\bf 0.353} & {\bf 0.993} & {\bf 0.989} & {\bf 0.020} & {\bf 0.767} & 51,043 & 101,733  \\
\bottomrule
\end{tabular}
\end{center}
\centering
\end{table*}

\end{document}